\documentclass{ieeetj}

\usepackage{cite}
\usepackage{amsmath,amssymb,amsfonts}
\usepackage{graphicx,color}
\usepackage{textcomp}
\usepackage{xcolor}
\usepackage{hyperref}
\hypersetup{hidelinks=true}
\usepackage{booktabs}
\usepackage{multirow}
\usepackage{array}
\usepackage{pifont}
\usepackage[flushleft]{threeparttable}
\usepackage{algorithm}
\usepackage[noend]{algpseudocode}
\usepackage[nolist,nohyperlinks]{acronym}
\usepackage[flushleft]{threeparttable}
\usepackage[caption=false,font=footnotesize]{subfig}
\usepackage[binary-units]{siunitx}
\usepackage{wasysym}

\newcommand{\cmark}{\checkmark}
\newcommand{\xmark}{\ding{53}}
\newcommand{\mypar}[1]{\textbf{#1}.}

\newtheorem{behavior}{Behavior}

\def\BibTeX{{\rm B\kern-.05em{\sc i\kern-.025em b}\kern-.08em
    T\kern-.1667em\lower.7ex\hbox{E}\kern-.125emX}}
\AtBeginDocument{\definecolor{tmlcncolor}{cmyk}{0.93,0.59,0.15,0.02}\definecolor{NavyBlue}{RGB}{0,86,125}}

\def\OJlogo{\vspace{-4pt}}
\def\seclogo{\vspace{10pt}}

\algnewcommand{\LineInput}[1]{\State \textbf{Input:} #1} 
\algnewcommand{\LineOutput}[1]{\State \textbf{Output:} #1}

\makeatletter
\def\@affil{%
\ifnum\affcount>\z@%
   \tempcount=\z@%
   \loop%
   \ifnum\affcount>\z@%
     \advance\tempcount\@ne%
         {\afffont\csname affil\romannumeral\the\tempcount\endcsname\par}%
    \advance\affcount\m@ne%
  \repeat%
\fi%
}
\makeatother
\DeclareMathAlphabet{\pazocal}{OMS}{zplm}{m}{n}

\newcommand{\Cs}{\pazocal{C}}
\newcommand{\Ds}{\pazocal{D}}

\newcommand{\Es}{\pazocal{E}}
\newcommand{\Ts}{\pazocal{T}}

\newcommand{\Vs}{\pazocal{V}}
\newcommand{\Fs}{\pazocal{F}}
\newcommand{\Ss}{\pazocal{S}}

\newcommand{\Js}{\pazocal{J}}
\newcommand{\Ms}{\pazocal{M}}

\newcommand{\Hs}{\pazocal{H}}

\newcommand{\pbf}{\mathbf{p}}

\newcommand{\xbf}{\mathbf{x}}

\newcommand{\Bbf}{\mathbf{B}}
\newcommand{\xibf}{\boldsymbol{\xi}}

\newcommand{\Gbb}{\mathbb{G}}

\newtheorem{definition}{Definition}
\newtheorem{problem}{Problem}
\newtheorem{objective}{Objective}

\acrodef{omniplanner}[OmniPlanner]{Universal exploration and inspection path planner}
\acrodef{fov}[FoV]{Field of View}
\acrodef{ve}[VE]{Volumetric Exploration}
\acrodef{vi}[VI]{Visual Inspection}
\acrodef{tr}[TR]{Target Reach}
\acrodef{esdf}[ESDF]{Euclidean Signed Distance Field}
\acrodef{sdf}[SDF]{Signed Distance Field}
\acrodef{tsp}[TSP]{Traveling Salesman Problem}
\acrodef{nbv}[NBV]{Next-Best-View}
\acrodef{mav}[MAV]{Micro Aerial Vehicle}
\acrodef{rov}[ROV]{Remote Operated Vehicle}
\acrodef{slam}[SLAM]{Simultaneous Localization and Mapping}
\acrodef{dtw}[DTW]{Dynamic Time Warping}
\acrodef{ar1}[AR-1]{Aerial Robot 1}
\acrodef{ar2}[AR-2]{Aerial Robot 2}
\acrodef{gr1}[GR-1]{Ground Robot 1}
\acrodef{ur1}[UR-1]{Underwater Robot 1}
\acrodef{am1}[AM-1]{Autonomy Module 1} 
\acrodef{am2}[AM-2]{Autonomy Module 2} 
\acrodef{am3}[AM-3]{Autonomy Module 3} 
\acrodef{am4}[AM-4]{Autonomy Module 4} 
\acrodef{am5}[AM-5]{Autonomy Module 5} 
\acrodef{c1}[C-1]{Configuration 1}
\acrodef{c2}[C-2]{Configuration 2}

\usepackage[nonumberlist,sort=use]{glossaries}
\makenoidxglossaries

\newglossaryentry{volume}{
    name=$V$, 
    description={Bounded environment volume}}

\newglossaryentry{embodiment morphology}{
    name=$R_{\mu}$, 
    description={The robot embodiment morphology, including its body form, sensors, and actuators }}

\newglossaryentry{embodiment constraints}{
    name=$C_{\mu}$, 
    description={The motion constraints related to a particular robot morphology }}

\newglossaryentry{robot configurations}{
    name=$\Xi$, 
    description={Set of collision-free robot configurations}}

\newglossaryentry{robot configuration}{
    name=$\xibf$, 
    description={Robot configuration $\xi = [p^x,p^y,p^z,\psi,\vartheta_a]\in\Xi$}}  

\newglossaryentry{pitch angle}{
    name=$\vartheta_a$, 
    description={Pitch angle of an actuated onboard sensor}}

\newglossaryentry{set of sensor modalities}{
    name=$\Ss$, 
    description={Set of onboard sensing modalities}}

\newglossaryentry{depth sensor}{
    name=$\Ds$, 
    description={Depth sensor with field of view $[F_H^{\Ds},F_V^{\Ds}]$ and maximum range $d_{\max}^{\Ds}$}}

\newglossaryentry{camera sensor}{
    name=$\Cs$, 
    description={Camera sensor with field of view $[F_H^{\Cs},F_V^{\Cs}]$ and maximum range $d_{\max}^{\Cs}$}}
    
\newglossaryentry{voxelized SDF grid}{
    name=$\Ms$, 
    description={Voxelised signed distance field grid}}

\newglossaryentry{voxel size}{
    name=$r_V$, 
    description={Fixed resolution of the volumetric map $\Ms$}}

\newglossaryentry{elevation map}{
    name=$\Hs$, 
    description={Elevation map with dimensions $[d_h^x,d_h^y]$}}

\newglossaryentry{elevation grid cell size}{
    name=$r_H$, 
    description={Fixed resolution of the elevation map $\Hs$}}

\newglossaryentry{voxel}{
    name=$m$, 
    description={Voxel in the volumetric map}}

\newglossaryentry{elevation map grid cell}{
    name=$h$, 
    description={Grid cell of the elevation map}}

\newglossaryentry{configurations observing voxel using depth}{
    name=$\Xi_m^{\Ds}$, 
    description={Set of collision-free configurations from which a voxel $m$ is observable by the depth sensor $\Ds$}}

\newglossaryentry{configurations observing voxel using camera}{
    name=$\Xi_m^{\Cs}$, 
    description={Set of collision-free configurations from which a voxel $m$ is observable by the camera sensor $\Cs$}}

\newglossaryentry{residual volume}{
    name=$V_{\mathrm{res}}$, 
    description={Residual volume}}

\newglossaryentry{residual surface}{
    name=$S_{\mathrm{res}}$, 
    description={Residual surface}}

\newglossaryentry{set of occupied voxels}{
    name=$\Ms_{\mathrm{occ}}$, 
    description={Set of occupied voxels in the volumetric map}}

\newglossaryentry{local graph}{
    name=$\Gbb_L$, 
    description={Local graph with vertex set $\Vs_L$ and edge set $\Es_L$}}

\newglossaryentry{local bounding box}{
    name=$\Bbf_L$, 
    description={Local bounding box}}

\newglossaryentry{current robot configuration}{
    name=$\xibf_0$, 
    description={Current robot configuration}}

\newglossaryentry{robot bounding box}{
    name=$\Bbf_R$, 
    description={Robot bounding box}}

\newglossaryentry{configurations within local bounding box}{
    name=$\Xi_{\Bbf_L}$, 
    description={Set of collision-free configurations within the local bounding box}}

\newglossaryentry{percentage of points sampled using Gaussian distribution}{
    name=$\eta$, 
    description={Percentage of points sampled using the Gaussian distribution}}

\newglossaryentry{robot configuration within local bounding box}{
    name=$\xibf_r$, 
    description={Robot configuration within the local bounding box}}  

\newglossaryentry{closest vertex in the graph}{
    name=$\nu_c$, 
    description={Closest vertex in the graph}}  

\newglossaryentry{radius from a vertex}{
    name=$e_{\max}$, 
    description={Radius from a vertex in the graph}}  

\newglossaryentry{new vertex}{
    name=$\nu_r$, 
    description={New vertex generated by moving the configuration $\xibf_r$ closer to $\nu_c$}}  

\newglossaryentry{maximum number of vertices}{
    name=$n_{\max}^{\Vs}$, 
    description={Maximum number of vertices allowed to connect to the graph}}  

\newglossaryentry{maximum number of edges}{
    name=$n_{\max}^{\Es}$, 
    description={Maximum number of edges allowed to connect to the graph}}  

\newglossaryentry{global grapah}{
    name=$\Gbb_G$, 
    description={Global graph with $\{\Vs_G,\Es_G\}$ vertices and edges sets}}

\newglossaryentry{current vertex of ith iteration}{
    name=$\nu_0^i$, 
    description={Current vertex of ith iteration}}

\newglossaryentry{sparse local graph}{
    name=$\Gbb_{L,\mathrm{sparse}}^i$, 
    description={Sparse local graph at ith iteration}}

\newglossaryentry{homing path}{
    name=$\sigma_{\mathrm{home}}$, 
    description={Collision-free return-to-home path}}

\newglossaryentry{home vertex}{
    name=$\nu_{\mathrm{home}}$, 
    description={Vertex at home location}}

\newglossaryentry{endurance of the robot}{
    name=$T_{\mathrm{thr}}$, 
    description={Robot's endurance}}

\newglossaryentry{nominal speed}{
    name=$v_{\mathrm{nom}}$, 
    description={Nominal commanded speed}}

\newglossaryentry{current time}{
    name=$t$, 
    description={Current time}}

\newglossaryentry{global path}{
    name=$\sigma_{G}$, 
    description={Global collision-free path}}

\newglossaryentry{target goal}{
    name=$\mathbf{p}_g$, 
    description={Target goal for local planner during global repositioning}}

\newglossaryentry{nominal clearance}{
    name=$h_0$, 
    description={Nominal clearance representing the robot's hight}}

\newglossaryentry{maximum slope}{
    name=$\theta_{\max}$, 
    description={Maximum allowable slope}}

\newglossaryentry{proximity threshold}{
    name=$d_{\max}$, 
    description={Proximity threshold}}

\newglossaryentry{free voxels}{
    name=$\Ms_{\mathrm{free}}$, 
    description={Set of free voxels in the volumetric map}}

\newglossaryentry{occupied volume}{
    name=$V_{\mathrm{occ}}$, 
    description={Occupied volume}}

\newglossaryentry{free volume}{
    name=$V_{\mathrm{free}}$, 
    description={Free volume}}

\newglossaryentry{straight line function}{
    name=$\gamma_{ij}$, 
    description={Straight line function}}

\newglossaryentry{volumetric information gain}{
    name=$\Gamma_{\mathrm{VE}}$, 
    description={Volumetric information gain}}

\newglossaryentry{local path}{
    name=$\sigma_{L}$, 
    description={Local candidate collision-free path}}

\newglossaryentry{set of candidate paths}{
    name=$\Sigma_{L}$, 
    description={Set of candidate collision-free paths}}

\newglossaryentry{local exploration objective}{
    name=$J_{\mathrm{VE}}^L$, 
    description={Local exploration objective function}}

\newglossaryentry{path length from vertex to vertex}{
    name=$\lambda_l$, 
    description={Path length from the root vertex $\nu_0$ to the vertex $\nu_k$ along $\Gbb_L$}}

\newglossaryentry{distance-based penalty factor}{
    name=$\mu_l$, 
    description={Distance-based penalty factor}}

\newglossaryentry{path direction penalty}{
    name=$\lambda_d$, 
    description={Path penalty for deviation from the current exploration direction}}

\newglossaryentry{direction-based penalty factor}{
    name=$\mu_d$, 
    description={Direction-based penalty factor}}

\newglossaryentry{best local path}{
    name=$\sigma_{L}^\star$, 
    description={Optimal local collision-free path}}

\newglossaryentry{frontiers set}{
    name=$\Fs$, 
    description={Set of frontiers in the global graph}}

\newglossaryentry{minimum frontier threshold}{
    name=$\Gamma_{\mathrm{thr},\Fs}$, 
    description={Threshold on volumetric information gain for a vertex to qualify as a frontier}}

\newglossaryentry{frontier vertex}{
    name=$\nu_{\Fs}$, 
    description={Frontier vertex}}

\newglossaryentry{current vertex in global graph}{
    name=$\nu_{0,G}$, 
    description={Vertex in the global graph that corresponds to the current configuration $\xibf_0$}}

\newglossaryentry{set of shortest paths to the frontiers}{
    name=$\Sigma_{G,\Fs}$, 
    description={Set of shortest paths from the current vertex to all frontiers}}

\newglossaryentry{set of shortest paths to home}{
    name=$\Sigma_{G,\mathrm{home}}$, 
    description={Set of shortest paths from all frontiers to home vertex}}

\newglossaryentry{global exploration objective}{
    name=$J_{\mathrm{VE}}^G$, 
    description={Global exploration objective function}}

\newglossaryentry{remaining time}{
    name=$\Ts$, 
    description={Estimated remaining exploration time}}

\newglossaryentry{target surface set}{
    name=$S_I$, 
    description={Target surface set}}

\newglossaryentry{voxels of the target surface}{
    name=$\Ms_I$, 
    description={Set of voxels associated with $S_I$}}

\newglossaryentry{inspection path}{
    name=$\sigma_I$, 
    description={Collision-free inspection path}}

\newglossaryentry{set of sampled viewpoints}{
    name=$P_I$, 
    description={Set of sampled points within $\Bbf_{\mathrm{VI}}$}}

\newglossaryentry{inspection bounding box}{
    name=$\Bbf_{\mathrm{VI}}$, 
    description={Sampling bounding box for inspection}}

\newglossaryentry{set of robot orientations}{
    name=$O_v$, 
    description={Set of robot orientation (and camera pitch if available) $\{\psi,\vartheta_a\}$}}

\newglossaryentry{set of viewpoint configurations}{
    name=$\Xi_v$, 
    description={Set of viewpoint configurations}}

\newglossaryentry{ith robot viewpoint configuration }{
    name=$\xibf_v^i$, 
    description={Robot configuration corresponding to $<\psi_v^i,\vartheta_{a,v}^i>~\in O_v$}}

\newglossaryentry{set of all viewpoint candidates}{
    name=$\Xi_I$, 
    description={Set of all viewpoint candidates}}

\newglossaryentry{coverage-optimal subset of viewpoints}{
    name=$\Xi_I^\star$, 
    description={Coverage-optimal subset of viewpoints}}

\newglossaryentry{viewpoint vertices}{
    name=$\Vs_I^\star$, 
    description={Viewpoint vertices}}

\newglossaryentry{optimal inspection path}{
    name=$\sigma_{I}^\star$, 
    description={Optimal collision-free inspection path}}

\newglossaryentry{path length between successive configurations}{
    name=$d_l$, 
    description={Path length between successive configurations}}

\newglossaryentry{User-defined target position}{
    name=$\pbf_t$, 
    description={User-defined target position}}
    
\newglossaryentry{initial robot configuration}{
    name=$\xibf_{\mathrm{init}}$, 
    description={Initial robot configuration}}

\newglossaryentry{target reach path}{
    name=$\sigma_{T}^\star$, 
    description={Optimal collision-free path for target reach}}

\newglossaryentry{Predefined tolerance radius around the target position}{
    name=$\rho_{\mathrm{reach}}$, 
    description={Predefined tolerance radius around the target position}}

\newglossaryentry{Best vertex in the global graph}{
    name=$\nu_{\mathrm{best}}$, 
    description={Closest (best) vertex in the global graph to the target point}}

\newglossaryentry{Guiding path towards the best vertex}{
    name=$\sigma_{\mathrm{guide}}$, 
    description={Guiding path towards the best vertex}}

\newglossaryentry{Closest vertex to the target point along the global graph}{
    name=$\nu_c$, 
    description={Closest vertex to the target point along the global graph}}

\newglossaryentry{Distance from the target point to the closest vertex in the global graph}{
    name=$\rho_t$, 
    description={Distance from the target point to the closest vertex in the global graph}}

\newglossaryentry{Shortest path length from current vertex to the frontier vertex}{
    name=$d_f^l$, 
    description={Shortest path length from $\nu_0$ to $\nu_f$}}

\newglossaryentry{Euclidean distance between the position of frontier vertex and the target point}{
    name=$d_f^u$, 
    description={Euclidean distance between the position of $\nu_f$ and $\pbf_t$}}

\newglossaryentry{Balancing term for the best vertex selection}{
    name=$\lambda_{\mathrm{bal}}$, 
    description={Balancing term for the best vertex selection}}

\newglossaryentry{Lookahead Point along the guiding path}{
    name=$\pbf_{\mathrm{lh}}$, 
    description={Lookahead point along the guiding path}}

\newglossaryentry{Path distance from the robot position to the lookahead point}{
    name=$\rho_{\mathrm{lh}}$, 
    description={Path distance from the robot position to the lookahead point}}

\newglossaryentry{Lookahead distance metric}{
    name=$\Gamma_{\mathrm{TR}}$, 
    description={Lookahead distance metric}}

\newglossaryentry{leaf configuration}{
    name=$\xibf_N$, 
    description={Leaf configuration}}

\newglossaryentry{target reach objective function}{
    name=$\Js_{\mathrm{TR}}$, 
    description={Target reach objective function}}

\begin{document}

    \markboth{\MakeUppercase{omniplanner: Universal Exploration and Inspection Path Planning Across Robot Morphologies}}{\MakeUppercase{Zacharia} {et al.}}

    \title{\acs{omniplanner}: Universal Exploration and Inspection Path Planning Across Robot Morphologies}
    \author{Angelos Zacharia, Graduate Student Member, IEEE, Mihir Dharmadhikari, Mohit Singh, Member, IEEE, and Kostas Alexis, Member, IEEE}
    \affil{Department of Engineering Cybernetics, Norwegian University of Science and Technology (NTNU), 7034 Trondheim, Norway}
    \corresp{Corresponding author: Angelos Zacharia (email: angelos.zacharia@ntnu.no).}
    \authornote{This work was supported in part by the Research Council of Norway under Grant NCEI (No. 357451) and in part by the European Commission under the Horizon Europe Programme through Grants SYNERGISE (No. 101121321), AUTOASSESS (No. 101120732), SPEAR (No. 101119774), and DIGIFOREST (No. 101070405).}

    \begin{abstract}
       Autonomous robotic systems are increasingly deployed for mapping, monitoring, and inspection in complex and unstructured environments. However, most existing path planning approaches remain domain-specific (i.e., in the air, on land, or at sea), limiting their scalability and cross-platform applicability. This article presents \acs{omniplanner}, a unified planning framework for autonomous exploration and inspection across aerial, ground, and underwater robots. The method integrates volumetric exploration and viewpoint-based inspection, alongside target reach behaviors within a single modular architecture, complemented by a platform abstraction layer that captures morphology-specific sensing, traversability and motion constraints. This enables the same planning strategy to generalize across distinct mobility domains with minimal retuning. The framework is validated through extensive simulation studies and field deployments in underground mines, industrial facilities, forests, submarine bunkers, and structured outdoor environments. Across these diverse scenarios, \acs{omniplanner} demonstrates robust performance, consistent cross-domain generalization, and improved exploration and inspection efficiency compared to representative state-of-the-art baselines. Videos presenting the \acs{omniplanner} framework and demonstrating its field deployments across aerial, ground, and underwater robots are available at \url{https://ntnu-arl.github.io/omniplanner}, and the source code is publicly available at \url{https://github.com/ntnu-arl/gbplanner_ros/tree/gbplanner3}.
    \end{abstract}
    
    
    \maketitle
    
    \begin{figure*}[t]
        \centering
        \includegraphics[clip, trim=0cm 0cm 0cm 0cm, width=1\linewidth]{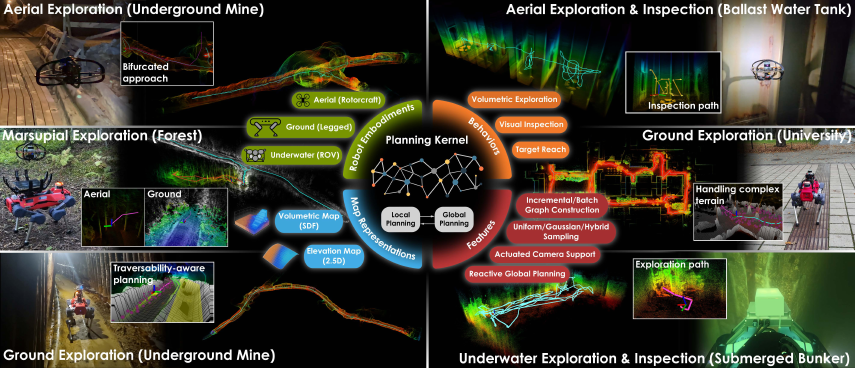}
        \caption{An overview of the \acs{omniplanner} core functionalities and features along with instances of field deployments. \acs{omniplanner} has been deployed across aerial, ground, and underwater robots in a diverse set of environments. Various aspects of the planner, including its planning behaviors, embodiment-specific adaptations, and bifurcated local-global architecture, are illustrated through examples of field deployments.}
        \label{fig:intro}
    \end{figure*}

    \section{INTRODUCTION}

    \IEEEPARstart{A}{utonomous} robotic operation in complex and unstructured environments requires the ability to actively acquire spatial information and systematically observe structures of interest. From underground mines \cite{scherer2022resilient,kottege2025heterogeneous} and industrial plants \cite{hutter2017towards} to subsea infrastructures \cite{jacobi2015autonomous} and disaster zones \cite{surmann2024lessons}, robots of diverse morphologies are increasingly deployed to perceive, map, and assess their surroundings, potentially without external supervision. These capabilities underpin applications such as search and rescue \cite{delmerico2019current}, infrastructure monitoring \cite{viswanathan2026xflie}, and environmental surveying \cite{popovic2020informative}, where human access is unsafe, impractical, or impossible.

    A large body of prior work has attempted to address autonomous planning for these tasks. Representative examples include volumetric exploration strategies for aerial robots \cite{zhou2021fuel,lindqvist2024tree}, traversability-aware planning for ground systems \cite{dixit2024step,lee2025trgplanner}, and coverage-oriented approaches for underwater inspection \cite{zacchini2022sensor, amer2025react}. While such methods achieve strong performance within their target domains, they are often tightly coupled to assumptions about the robot morphology, including the associated vehicle dynamics, sensing modalities, and environmental structure. Furthermore, they are typically optimizing for a single objective, such as exploration, inspection or target reach. Consequently, adapting these planners to new robot types or tasks typically requires substantial redesign, reparameterization, or parallel development of separate planning pipelines. Moreover, exploration and inspection are commonly treated as distinct problems, with limited support for transitioning between them within a unified planning architecture. This fragmentation restricts scalability and limits the transfer of autonomy across diverse robotic platforms.

    Despite the apparent diversity of robotic morphologies and domains, many of the underlying planning requirements remain shared. Aerial and underwater robots are both floating-base platforms operating in 3D space. Ground platforms, despite other morphological differences, must simultaneously reason about both obstacle avoidance and traversability over uneven terrain and complex geometries. Across morphologies, robots must repeatedly solve a common set of problems: selecting collision-free motions, reasoning over partially observed environments, and choosing viewpoints that maximize task-relevant information. These shared objectives suggest that autonomy across domains --at least for certain morphology classes such as those that can be approximated as (near-)holonomic systems-- need not rely on fundamentally different planners, but rather on a unified planning core that can be specialized through modular interfaces.
    
    Motivated by these observations, this paper introduces \acs{omniplanner}, a unified planning framework centered around a domain-agnostic planning kernel. The proposed framework decouples the core planning logic from robot-specific constraints, map representations, and task objectives. The planning kernel serves as a shared backbone for global and local planning, while behaviors such as volumetric exploration, visual inspection, and target reach are realized through modular objective functions. Robot-specific characteristics --including those of aerial, ground, and underwater embodiments-- are incorporated through lightweight adaptation layers, enabling the same planning kernel to be reused across platforms with minimal domain-specific tuning.

    The main contributions of this work are summarized as follows:
    \begin{itemize}
        \item \textbf{Planning Kernel architecture}: A unified, domain-agnostic planning kernel that serves as a shared core for global and local planning behaviors across heterogeneous robotic platforms that otherwise can be approximated as (near-)holonomic systems. 
        \item \textbf{Modular behavior integration}: A behavior abstraction layer that unifies volumetric exploration, visual inspection, and target reach within a unified planning framework and common planning kernel, enabling a multitude of tasks and the autonomous switching between behaviors without external intervention.
        \item \textbf{Cross-domain validation}: Extensive simulation studies benchmark \acs{omniplanner} against state-of-the-art methods across exploration, inspection and target reach behaviors, while experimental field deployments on aerial, ground, and underwater robots demonstrate its effectiveness across diverse environments, including underground mines, forests, submarine bunkers, industrial facilities, and structured outdoor settings.        
    \end{itemize}
    The above position \acs{omniplanner} in a distinct category compared to other methods that are limited in their ability to generalize regarding the robot morphologies they can guide and the operational domains (air, land, sea) in which they can successfully operate. The implementation of the method, alongside the environments used for evaluations in simulation and the datasets from field testing are available at \url{https://github.com/ntnu-arl/gbplanner_ros/tree/gbplanner3}.

    The remainder of this paper is organized as follows. Section~\ref{sec:related_work} reviews related work on exploration, inspection, and target reach planning approaches. Section~\ref{sec:problem_statement} formulates the planning problem and introduces the abstraction used to represent heterogeneous robots and environments. Section~\ref{sec:proposed_approach} details the proposed planning kernel and its associated adaptation layers. Section~\ref{sec:simulation_studies} presents simulation-based evaluations, while Section~\ref{sec:field_experiments} reports the results of extensive field experiments conducted with aerial, ground, and underwater robots. Finally, Section~\ref{sec:discussion} discusses lessons learned and the benefits of a unified planning framework, and Section~\ref{sec:conclusion} concludes the paper and discusses directions for future research.
  
    
    \section{RELATED WORK}
    \label{sec:related_work}

    This section outlines relevant literature, organized around the exploration, inspection, and target reach behaviors. 
    \subsection{Exploration Planning}
        Autonomous exploration has been extensively studied using two dominant approaches: frontier-based exploration and \ac{nbv} planning. Frontier-based methods select goals on the boundary between known free space and unknown space, originating from the seminal work in~\cite{yamauchi1997frontier} and later extended to multi-robot exploration~\cite{yamauchi1998frontier}. More recent frontier formulations have focused on enabling rapid exploration for agile aerial robots, e.g., by designing frontier selection strategies that support high-speed flight~\cite{cieslewski2017rapid}. Several studies have also compared frontier-based variants to highlight their trade-offs across representative environments and deployment conditions~\cite{jain2017comparative}. Exploiting implicit grouping of frontier voxels, the work in~\cite{dai2020fast} improves computational performance.

        \acs{nbv} planning instead chooses sensing configurations by optimizing an information objective, and has roots in early work on determining the next best view~\cite{connolly1985determination}. \acs{nbv} principles are broadly applicable across domains, including underwater perception, where view selection must respect sensing and visibility constraints~\cite{sheinin2016next}. Modern \acs{nbv} systems often adopt receding-horizon formulations, such as NBVP~\cite{bircher2016receding}, and have been accelerated through sampling-based improvements tailored to \acp{mav}~\cite{respall2021fast}. A well-known limitation of purely local \acs{nbv} selection is susceptibility to local minima, where a planner exhausts locally informative viewpoints and fails to relocate to distant informative regions. This limitation is explicitly discussed and addressed in large-scale 3D exploration settings~\cite{selin2019efficient}. Simultaneously, approaches such as ~\cite{bircher2016receding} that rely on sampling-based methods but build a single tree face computational challenges as the scale of the environment increases. Complementary approaches have explored motion-primitive libraries to enable fast, dynamically feasible exploration behaviors~\cite{dharmadhikari2020motion}. Since many exploration pipelines rely on sampling-based planning, improving sampling efficiency remains an active area, including methods that reduce computational overhead for online informative planning in unknown environments~\cite{schmid2020efficient}. Recent work has also proposed the ERRT framework, a tree-based next-best-trajectory formulation for 3D UAV exploration, which explicitly optimizes informative motion over a branching trajectory tree while maintaining real-time feasibility~\cite{lindqvist2024tree}.
        
        To mitigate local-minima behavior and improve scalability, hierarchical and integrated exploration planners combine local planning with global reasoning. Representative local--global frameworks include GBPlanner 2.0~\cite{kulkarni2022autonomous}, TARE~\cite{cao2021tare}, DSVP~\cite{zhu2021dsvp}, and FUEL~\cite{zhou2021fuel}, which typically maintain global structures (e.g., graphs or frontier sets) to support long-horizon repositioning while using local planners for collision-free execution. More recent works have further targeted global optimality and robustness, e.g., by incorporating frontier-omission awareness and altitude-stratified planning~\cite{zhang2023go}, or by exploiting submap structures to maintain exploration progress under severe odometry drift~\cite{schmid2021unified}. In parallel, information-theoretic exploration~\cite{tabib2021autonomous} has advanced through objectives based on mutual information and its tractable approximations, including Bayesian optimization for informative view selection~\cite{bai2016information}, state-lattice planning with information measures for subterranean environments~\cite{tabib2016computationally}, and real-time information-theoretic exploration using Gaussian mixture model maps~\cite{tabib2019real}. Exploration in dynamic environments has also been studied, e.g., by leveraging roadmap-style representations that enable efficient re-querying as the environment changes~\cite{tabib2019real}. Finally, learning-based exploration has gained traction, including imitation learning approaches derived from expert behavior in subterranean settings~\cite{reinhart2020learning} and broader learning-based formulations for adaptive informative path planning~\cite{popovic2024learning}. At the systems level, MAexp provides a generic high-efficiency platform for RL-based multi-agent exploration, combining continuous point-cloud environments, multiple MARL algorithms, and faster sampling to support more reproducible cross-scenario evaluation and improved sim-to-real fidelity~\cite{zhu2024maexp}.

        A related thread is uncertainty-aware exploration and active \ac{slam}, where the objective is not only map coverage but also improved localization and state estimation. Early active \acs{slam} work framed viewpoint selection through model predictive control and attractor-based exploration~\cite{leung2006active}, while later approaches incorporated information measures such as Kullback--Leibler divergence to guide exploration under particle-filter \acs{slam}~\cite{carlone2010application}. Active vision has also been used to improve localization quality via controllable sensing, for example through an active stereo head~\cite{davison2002simultaneous}. Uncertainty-aware planners such as RHEMPlanner explicitly consider estimation uncertainty during exploration and mapping~\cite{papachristos2017uncertainty}, and recent surveys summarize the broader active \acs{slam} landscape and open challenges~\cite{placed2023survey}.

    \subsection{Inspection and Coverage Planning}
        Inspection and coverage planning differ from exploration in that the objective is typically to observe a known or partially known target surface under sensing constraints, rather than only to expand the free-space map. Classical 3D coverage path planning methods usually assume a prior model and decompose the problem into viewpoint generation and route optimization. An early sampling-based formulation proposed in~\cite{englot2012sampling} addresses full-surface inspection in cluttered, occluded environments and provides probabilistic completeness guarantees for coverage planning. A representative online extension is the receding-horizon framework in~\cite{bircher2018receding}, which offers volumetric exploration and surface inspection, albeit not integrated within a single autonomous mission. Analogous to~\cite{bircher2016receding}, the method faces computational challenges in spatially extended missions as it samples a single random tree. In a more inspection-specific setting, ASSCPP uses an existing 3D reference model and sensor noise models to adaptively sample viewpoints toward low-coverage and low-accuracy regions, thereby improving both path efficiency and expected model quality~\cite{almadhoun2018coverage}.

        A major challenge in aerial inspection is scalability in large and cluttered 3D scenes. HCPP addresses this through a hierarchical decomposition that partitions the environment into subspaces, computes a global traversal order, and then solves local coverage paths within each subspace~\cite{cao2020hierarchical}. More recently, FC-Planner improves this idea through skeleton-guided space decomposition and specialized viewpoint generation, reducing redundant sampling and yielding faster coverage planning in complex scenes~\cite{feng2024fc}. Beyond geometric coverage alone, recent work has also emphasized visibility and reconstruction quality. Star-convex visibility planning constrains the trajectory to remain within safe-and-visible regions during inspection~\cite{liu2022star}, while GS-Planner uses 3D Gaussian Splatting to evaluate reconstruction completeness together with geometric and textural quality online, enabling quality-aware active reconstruction~\cite{jin2024gsplanner}.

        Another recent trend is to unify coverage and exploration for online modeling of unknown structures. SEAC departs from the conventional explore-then-exploit pipeline by jointly optimizing local coverage of low-quality surfaces and global exploration of unseen regions within a hierarchical framework, improving both reconstruction quality and efficiency~\cite{zhang2025seac}. Practical deployment has also motivated model-informed and cooperative variants, including BIM-supported path planning for building exterior inspection and multi-robot systems for 3D surface reconstruction~\cite{huang2023bim,hardouin2023multirobot}. At the evaluation level, CARIC highlights the growing importance of realistic benchmarking for inspection planners, especially in multi-UAV settings, by emphasizing not only completeness and efficiency but also inspection quality under practical constraints such as heterogeneous sensing and communication limits~\cite{cao2025cooperative}. Overall, the literature shows a clear shift from offline model-based coverage toward scalable, visibility-aware, and quality-driven online inspection planning.

    \subsection{Target Reach Planning}
        Target reach planning considers the problem of navigating a robot to a specified goal as quickly and safely as possible, typically in partially known or unknown environments. Recent work has increasingly adopted integrated global--local formulations to balance long-horizon route selection with fast local replanning. FAR Planner is representative of this direction, which incrementally builds a polygonal map and dynamically updates a visibility graph, enabling low-latency ``attemptable'' routing toward a goal while adapting to newly observed obstacles and dead ends~\cite{yang2022far}. For aerial robots, FASTER combines global guidance with local trajectory optimization and explicitly maintains a safe backup trajectory in known free space while planning a faster exploratory trajectory toward the goal, improving speed without sacrificing safety~\cite{tordesillas2019faster}.

        A closely related line of work targets aggressive goal-reaching in cluttered unknown environments. Bubble Planner improves high-speed local replanning through overlapping sphere corridors and a receding-horizon corridor reuse strategy, increasing replanning success and enabling smooth, dynamically feasible flight~\cite{ren2022bubble}. More recently, SUPER extends this safety-assured paradigm by planning directly on LiDAR point clouds and using differentiable trajectory optimization, achieving high-speed and robust waypoint navigation in complex unknown environments~\cite{ren2025safety}. In parallel, perception-driven local planners reduce reliance on explicit mapping by reasoning directly from onboard sensing. For example, depth-conditioned N-MPC embeds a learned collision model into receding-horizon control for real-time obstacle avoidance during waypoint reaching~\cite{jacquet2024n}, while reinforcement learning with deep collision encoding maps compressed depth observations, robot state, and goal information directly to low-latency control commands~\cite{kulkarni2024reinforcement}. Overall, recent target reach methods have shifted toward integrated global--local and perception-aware formulations that better trade off speed, safety, and online adaptability.\\

    Despite the strong performance of recent exploration, inspection, and target reach planners, most remain specialized either to a specific task or to a particular robot embodiment category and sensing configuration. As a result, transferring these approaches across tasks and robot morphologies often requires substantial redesign, reparameterization, or even the development of separate planning pipelines. Among existing local-global planning frameworks, GBPlanner~2.0~\cite{kulkarni2022autonomous} represents the closest architectural relative to \acs{omniplanner}, as both employ a bifurcated local-global planning strategy. However, important differences exist. While GBPlanner~2.0 is designed exclusively for exploration, \acs{omniplanner} is a multi-behavior solution supporting (a) volumetric exploration, (b) visual inspection, and (c) target reach behaviors through the instantiation of modular objective functions on a shared planning kernel. Furthermore, \acs{omniplanner} decouples the planning logic from robot embodiment category through lightweight adaptation layers, enabling deployment across aerial, ground, and underwater robots without modifying the underlying architecture. At the planning kernel level, \acs{omniplanner} extends graph generation through the introduction of Gaussian and hybrid sampling strategies in addition to uniform sampling, and incorporates a batch graph-construction approach that improves graph connectivity discovery in large-scale environments. Simultaneously, the method supports active cameras on robots. \acs{omniplanner} is demonstrated in a set of comprehensive studies and presents the distinctive feature of verification across aerial, ground, and underwater systems. Collectively, these capabilities enable a unified, domain-agnostic planning framework that supports multiple autonomous behaviors across heterogeneous robotic platforms while preserving a common underlying planning architecture.

     \section{PROBLEM STATEMENT}
    \label{sec:problem_statement}

    This work considers autonomous path planning under partial observability, where information about the environment is acquired through motion while respecting embodiment-specific motion and sensing constraints. The problem is formulated in a domain-agnostic manner to support a unified planning kernel that can be instantiated across heterogeneous robotic platforms and task objectives.
    
    Let $V \subset \mathbb{R}^3$ denote a bounded environment volume. The robotic platform is characterized by its embodiment morphology $R_{\mu}$ and associated motion constraints $C_{\mu}$, which define the set of collision-free configurations $\Xi$. A configuration $\xibf \in \Xi$ is defined as the robot position $[p^x,p^y,p^z]$ and yaw angle $\psi$, and when available, further includes a single rotational degree of freedom corresponding to the pitch angle $\vartheta_a$ of an actuated onboard sensor ($\xibf=[p^x,p^y,p^z,\psi]$ or $\xibf=[p^x,p^y,p^z,\psi,\vartheta_a]$). In this work, a set of onboard sensing modalities $\Ss = \{\Ds,\Cs\}$, corresponding to a depth and a camera sensor (possibly but not necessarily realized on the same device), respectively, is characterized by bounded \acp{fov}, finite sensing range, and configuration-dependent visibility constraints. These properties induce geometric observability relations between robot configurations and environment regions.

    The environment is represented by a spatial map that combines a voxelized \ac{sdf} grid $\mathcal{M}$ (also referred to as the volumetric map) with fixed resolution $r_V$ and (optionally) a $2.5$D grid-based elevation map $\mathcal{H}$ (for ground robots) with fixed resolution $r_H$. Each voxel $m \in \mathcal{M}$ encodes the belief state of the corresponding spatial region as free, occupied, or unknown, as well as the distance to the closest surface (referred to as \ac{sdf} distance). The function $\text{SDF}(\xbf)$ returns the \ac{sdf} distance of the voxel in which $\xbf \in \mathbb{R}^3$ lies. This representation supports collision checking, visibility reasoning, and information-theoretic evaluation within the planning process. The elevation map $\Hs$ is implemented as a $2.5$D sliding-window map of dimensions $[d_h^x, d_h^y]$, centered at the current robot location. Each grid cell $h \in \Hs$ stores the estimated ground elevation at the corresponding $[x,y]$ coordinate. This representation enables traversability-aware planning for ground robots.

    Due to the inherent limitations of range-based and view-constrained sensing, which primarily observe surface boundaries and are subject to occlusions, certain regions of the environment may remain fundamentally unobservable. Let $\Xi_m^{\mathcal{D}} \subset \Xi$ denote the set of collision-free configurations from which a voxel $m$ is observable by the depth sensor $\mathcal{D}$. Similarly, let $\Xi_m^{\mathcal{C}} \subset \Xi$ denote the set of configurations from which an occupied voxel $m$ is observable by the camera sensor $\mathcal{C}$. Definitions \ref{def:residual_volume} and \ref{def:residual_surface} capture intrinsic limits of environment observability imposed by the robot’s embodiment and sensing modalities.
    \begin{definition}[Residual Volume]
        The residual volume $V_{\mathrm{res}}\subset V$ is defined as the subset of the environment volume consisting of voxels that cannot be observed by the depth sensor from any collision-free configuration:
        \begin{equation}
            V_{\mathrm{res}} = \bigcup_{m \in \mathcal{M}} \left( m \mid \Xi_m^{\mathcal{D}} = \emptyset \right).
        \end{equation}
        \label{def:residual_volume}
    \end{definition}
    \begin{definition}[Residual Surface]
        Let $\mathcal{M}_{\mathrm{occ}} \subset \mathcal{M}$ denote the set of occupied voxels corresponding to observed surfaces. The residual surface $S_{\mathrm{res}}$ is defined as the subset of occupied voxels that cannot be observed by the camera from any collision-free configuration:
        \begin{equation}
            S_{\mathrm{res}} = \bigcup_{m \in \mathcal{M}_{\mathrm{occ}}} \left( m \mid \Xi_m^{\mathcal{C}} = \emptyset \right).
        \end{equation}
        \label{def:residual_surface}
    \end{definition}
    
    Based on the above definitions, the planning problem addressed in this work is overarchingly formulated at the kernel level, independently of any specific task or behavior.
    \begin{problem}[Overarching Planning Problem] 
        Given a bounded environment volume $V$, a robot with configuration space $\Xi$ and motion constraints $C_{\mu}$, and sensing modalities $\mathcal{S}$, determine a collision-free trajectory $\sigma$ that respects all motion and sensing constraints and optimizes an extrinsic objective $\mathcal{J}$ over the environment. The objective is either target reach or an information-gathering task, specifically exploration or inspection. For the latter two cases, the objective evaluates how effectively the robot’s trajectory acquires task-relevant information through sensing, based on a volumetric map representation $\mathcal{M}$.
    \end{problem}

    \begin{objective}[Exploration]
    As the exploration objective, the method considers the planning of a path and viewpoints to unveil all possible volume within a defined bounded box, given no prior information and subject to the considered sensing and motion model. 
    \end{objective}

    \begin{objective}[Inspection]
    As the inspection objective, the method considers the planning of a path and viewpoints to enable the coverage of all possible surfaces within a defined bounded box, given a representation of the underlying map (possibly through the exploration step). 
    \end{objective}

    \begin{figure*}[t!]
        \centering
        \includegraphics[clip, trim = 0cm 0cm 0cm 0cm, width=1\linewidth]{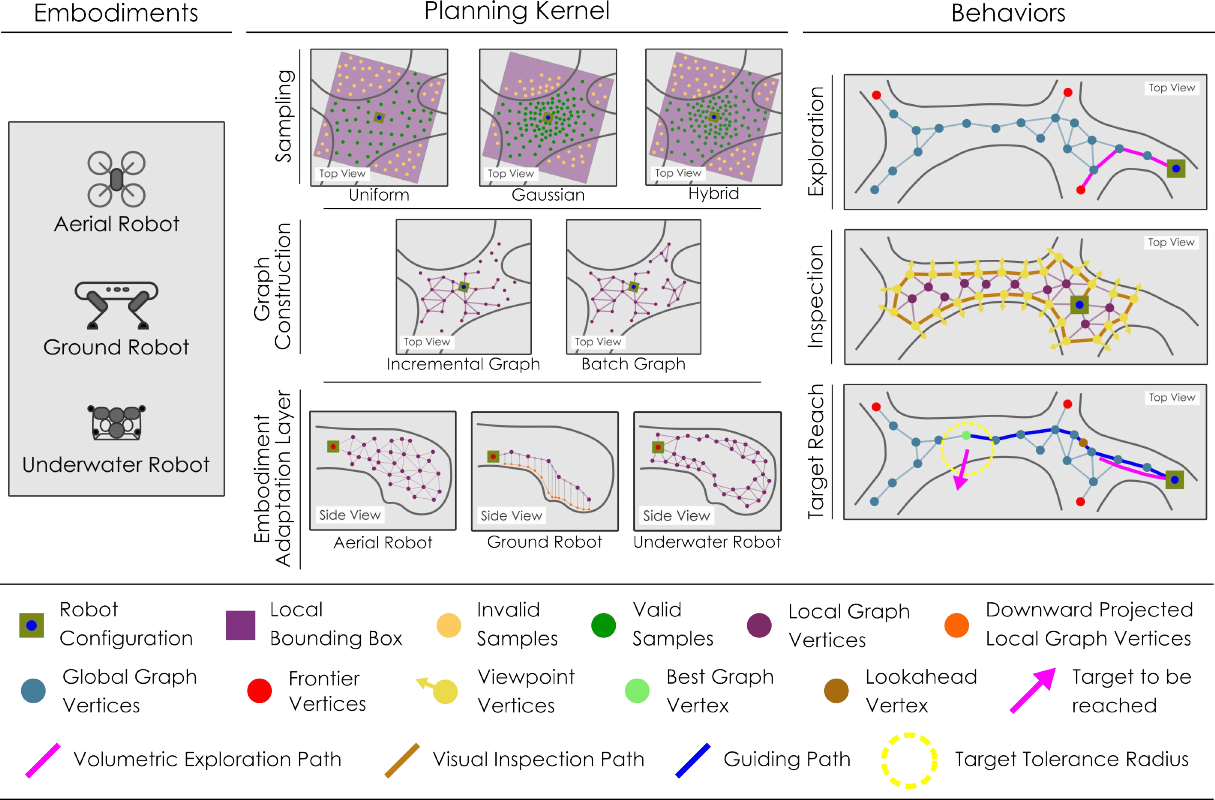}
        \caption{\acs{omniplanner} -- a platform-agnostic planning kernel supports global and local planning, while adaptation layers abstract robot embodiments and map representations. Task-specific behaviors are instantiated as objectives and features on top of the shared kernel, enabling reusable autonomy across heterogeneous platforms.}
        \label{fig:overview}
    \end{figure*}

    \begin{objective}[Target Reach]
    As the target reach objective, the method considers the planning of a path to reach a user-defined target destination, with or without any prior map information and subject to the considered sensing and motion model.  
    \end{objective}

    Subsequently, we present how \acs{omniplanner} addresses the considered problem and gives rise to task-specific autonomous behaviors through the instantiation of different objective functions, constraints, and termination conditions, while preserving the same underlying planning architecture. Formal definitions for each behavior are provided. Definitions 1–2, Problem 1, and Objectives 1–3 establish a common mathematical formulation of observability, motion constraints, and task objectives that is independent of robot morphology and mission type. This formulation serves as the foundation for realizing exploration, inspection, and target reach behaviors within a single architecture across aerial, ground, and underwater robotic platforms.

    
     \section{PROPOSED APPROACH}
    \label{sec:proposed_approach}

    \begin{algorithm*}[t]
        \begingroup
        \caption{\acs{omniplanner} -- Unified Planning Framework}
        \label{alg:omniplanner}
        \begin{algorithmic}[1]
            \State $R_{\mu}, C_{\mu} \gets \textbf{GetRobotAbstraction}()$ 
            \Comment{Embodiment adaptation layer}
            \State $\mathcal{S}=\{\mathcal{D},\mathcal{C}\} \gets \textbf{GetSensorModel}()$ 
            \Comment{Sensor abstraction layer}
            \State $\mathcal{J} \gets \textbf{GetMissionObjective}()$ 
            \Comment{Mission abstraction layer}
            \State $\mathbb{G}_G=\{\mathcal{V}_G,\mathcal{E}_G\} \gets \emptyset$
            \While{$\textbf{not } \big(\textbf{TaskComplete}(\mathcal{J})$ \textbf{or} $\textbf{EnduranceCritical}()\big)$}
                \State $\boldsymbol{\xi}_0,\mathcal{M},\mathcal{H} \gets \textbf{UpdateStateAndMap}()$
                \State $\mathbb{G}_L=\{\mathcal{V}_L,\mathcal{E}_L\} 
                \gets \textbf{BuildLocalGraph}(\boldsymbol{\xi}_0,R_{\mu},C_{\mu},\mathcal{M},\mathcal{H})$
                \Comment{Planning kernel}
                \State $\mathbb{G}_G 
                \gets \textbf{UpdateGlobalGraph}(R_{\mu},C_{\mu},\mathbb{G}_L,\mathbb{G}_G)$
                \Comment{Planning kernel}
                \If{$\mathcal{J}=\mathcal{J}_{\mathrm{VE}}$}
                    \State $\sigma^\star \gets 
                    \textbf{VolumetricExploration}(\boldsymbol{\xi}_0,R_{\mu},C_{\mu},\mathbb{G}_L,\mathbb{G}_G,\mathcal{M},\mathcal{D})$ \Comment{VE Behavior (Algorithm~\ref{alg:exploration})}
                \ElsIf{$\mathcal{J}=\mathcal{J}_{\mathrm{VI}}$}
                    \State $\sigma^\star \gets 
                    \textbf{VisualInspection}(\boldsymbol{\xi}_0,R_{\mu},C_{\mu},\mathbb{G}_L,\mathcal{M},\mathcal{C})$ \Comment{VI Behavior (Algorithm~\ref{alg:inspection})}
                \ElsIf{$\mathcal{J}=\mathcal{J}_{\mathrm{TR}}$}
                    \State $\sigma^\star \gets 
                    \textbf{TargetReach}(\boldsymbol{\xi}_0,\mathbf{p}_t,R_{\mu},C_{\mu},\mathbb{G}_L,\mathbb{G}_G,\mathcal{M})$ \Comment{TR Behavior (Algorithm~\ref{alg:target_reach})}
                \EndIf
                \State $\textbf{ExecutePath}(\sigma^\star)$
            \EndWhile
            \State $\sigma_{\mathrm{home}} \gets \textbf{ReturnToHome}(\boldsymbol{\xi}_0,R_{\mu},C_{\mu},\mathbb{G}_G)$
            \State $\textbf{ExecutePath}(\sigma_{\mathrm{home}})$
        \end{algorithmic}
        \endgroup
    \end{algorithm*}

    This section presents \acs{omniplanner}, a unified planning framework structured around a domain- and morphology-agnostic planning kernel. The kernel provides a shared backbone for path planning across heterogeneous robotic platforms, while autonomous behaviors are realized through modular objective functions and feature extensions layered on top of this core, as shown in Fig.~\ref{fig:overview}. Algorithm~\ref{alg:omniplanner} summarizes the overall planning framework and shows the interaction between the embodiment adaptation layer, the sensor abstraction layer, the mission abstraction layer, and the planning kernel. The embodiment adaptation layer defines the robot morphology $R_{\mu}$ and the corresponding motion constraints $C_{\mu}$, while the sensor abstraction layer provides the sensing model $\mathcal{S}=\{\mathcal{D},\mathcal{C}\}$. The planning kernel constructs and updates the local and global graphs, $\Gbb_L$ and $\Gbb_G$, using these shared representations. The objective-dependent step is the selection of $\mathcal{J}$, which determines whether the common kernel is used for volumetric exploration, visual inspection, or target reach behavior.

    \subsection{Planning Kernel}
        \label{sec:planning_kernel}
        The planning kernel operates on the robot configuration space $\Xi$ and an incrementally constructed volumetric map $\mathcal{M}$ and elevation map $\Hs$, independent of task specification and robot embodiment. It adopts a bifurcated planning structure, inspired by our previous work \cite{kulkarni2022autonomous}, composed of tightly coupled Local and Global Planning Modules that enable scalable planning in large-scale three-dimensional environments.
    
        \subsubsection{\textbf{Local Planning Module}}
            The Local Planning Module constructs a bounded, sampling-based dense graph $\Gbb_L$ in a box of dimensions $\Bbf_L = [b_L^x, b_L^y, b_L^z]$ around the robot’s current configuration $\xibf_0$. The purpose of this graph is to represent the locally reachable subset of the configuration space $\Xi$ under embodiment-specific motion and sensing constraints, while maintaining bounded computational complexity.
           
            Given the volumetric map $\Ms$, elevation map $\Hs$, and a bounding box $\Bbf_R = [b_R^x, b_R^y, b_R^z]$ encoding the robot’s physical extent, a set $\Xi_{\Bbf_L}$ of collision-free configurations within $\Bbf_L$ is randomly sampled. In contrast to previous graph-based planners that rely exclusively on uniform sampling~\cite{kulkarni2022autonomous,dang2020graph}, \acs{omniplanner} supports three different sampling distributions to better balance local connectivity and long-range reachability across diverse operating conditions:
            \begin{itemize}
                \item \textit{Uniform}: Uniform distribution along each axis within $\Bbf_L$. This is the most generic distribution that can be used to enable efficient planning in a wide variety of environments.
                \item \textit{Gaussian}: Gaussian distribution centered at $\xibf_0$ with a user-defined covariance $\mathbf{\Lambda}$. This distribution can enable improved reachability when operating in narrow environments, as the planner samples densely around the robot, at the cost of worse reach in the volume further away.
                \item \textit{Hybrid}: Combination of \textit{Uniform} and \textit{Gaussian}. In this distribution, $\eta~\%$ samples are sampled using the \textit{Gaussian} distribution and the rest using \textit{Uniform}. This balances operation in narrow environments with the broader coverage provided by the uniform sampling.
            \end{itemize}
            The samples in $\Xi_{\Bbf_L}$ are connected by admissible edges to form the local graph $\Gbb_L$ along with its vertex and edge sets $\Vs_L$, $\Es_L$ respectively. An edge is admissible if it lies entirely in the collision-free part of the volumetric map $\Ms_{\mathrm{free}} \subset \Ms$ and respects the robot motion constraints $C_{\mu}$ (further details presented in Section~\ref{subsec:adapt}). The planner supports two strategies for building $\Gbb_L$ using $\Xi_{\Bbf_L}$:
            \begin{itemize}
                \item \textit{Incremental}: Analogous to \cite{kulkarni2022autonomous,dang2020graph}, in this method, one robot configuration is sampled and added to $\Gbb_L$ at a time. Specifically, a random sample $\xibf_r$ is sampled inside $\Bbf_L$ using the selected sampling distribution. The closest vertex $\nu_c \in \Vs_L$ to $\xibf_r$ is selected, and if $\xibf_r$ is further than the maximum allowed edge length $e_{\max}$, $\xibf_r$ is moved closer to $\nu_c$ along the line joining them to create the new vertex $\nu_r$. Next, the vertices $\{ \nu_{nb} \} \in \Vs_L$ within a radius $e_{\max}$ of $\nu_r$ are connected if the straight line edges are admissible. The process is then repeated until the number of vertices or edges in $\Gbb_L$ reaches the user-defined limits $n^{\Vs}_{\max},~ n^{\Es}_{\max}$.
                \item \textit{Batch}: Unlike the incremental graph-construction strategy employed by GBPlanner~2.0~\cite{kulkarni2022autonomous} and related approaches~\cite{dang2020graph}, the \textit{Batch} approach samples a batch of $n^{\Vs}_{\max}$ configurations at a time. The vertices within a radius of $e_{\max}$ of each other are connected if admissible straight line edges exist. The parts of $\Gbb_L$ disconnected from the vertex $\nu_0$ corresponding to $\xibf_0$ are pruned. This process is then repeated until the number of vertices or edges in $\Gbb_L$ reaches the user-defined limits $n^{\Vs}_{\max},~ n^{\Es}_{\max}$.
            \end{itemize}
            
        \subsubsection{\textbf{Global Planning Module}}
            The Global Planning Module maintains a sparse, incrementally constructed graph $\mathbb{G}_G = \{\mathcal{V}_G,\mathcal{E}_G\}$ that captures the connectivity of the mapped configuration space over time. In contrast to the local planning graph, which is transient and restricted to a bounded planning volume, the global graph persists across planning iterations and grows as the robot moves through previously unmapped regions of the environment.

            In each local planning iteration $i$, shortest paths $\{ \sigma \}$ in the graph $\Gbb_L^i$ are calculated using Dijkstra's algorithm \cite{dijkstra2022note} from $\nu_0^i$ to each vertex $\nu \in \Vs_L^i$ ($(\cdot)^i$ represents the variable corresponding to iteration $i$). To limit redundancy and control the growth of the global graph, these paths are clustered based on a similarity metric derived from \ac{dtw}~\cite{bachrach2013trajectory}. Let $\mathcal{R} = \{R_k\}$ denote the set of cluster representative paths. Given two paths $\sigma_i,\sigma_j\in\{\sigma\}$, their similarity is quantified through a normalized distance $\delta(\sigma_i,\sigma_j) = \frac{\mathrm{DTW}(\sigma_i,\sigma_j)}{L^2}$, where $L = \min(\mathrm{len}(\sigma_i), \mathrm{len}(\sigma_j))$. Clustering is performed using a greedy assignment strategy with threshold $\tau_c$, where paths are processed in decreasing order of length and assigned to the first representative $R_k \in \mathcal{R}$ satisfying $\delta(\sigma_i, R_k) \leq \tau_c$, or initialize a new representative otherwise. Representatives with length below a threshold $\ell_{\min}$ are removed, after which all paths are reassigned to the nearest remaining representative. The resulting representatives define a sparse subset $\Gbb_{L, \mathrm{sparse}}^i \subset \Gbb_L^i$ that spans $\Bbf_L$. Each vertex and edge from $\Gbb_{L, \mathrm{sparse}}^i$ is added to $\Gbb_G$ such that $\Gbb_G = \Gbb_G \bigcup \Gbb_{L, \mathrm{sparse}}^i$. Each vertex in $\Gbb_{L, \mathrm{sparse}}^i$ is connected to vertices in $\Gbb_G$ within a radius $e_{\max}$. 
            
            Through this incremental aggregation of locally validated motion structure, the resulting graph remains lightweight while providing a meaningful approximation of the traversable configuration space. It enables efficient long-horizon path queries between arbitrary previously visited configurations, while maintaining bounded memory usage and computational complexity without requiring dense sampling of the entire configuration space. Furthermore, the Global Planning Module keeps track of the robot's endurance to provide safe return to home functionality. In each local planning iteration, the global planner calculates a path $\sigma_{\mathrm{home}}$ from the current robot location to the start location $\nu_{\mathrm{home}} \in \Vs_G$, along $\Gbb_G$. If $\mathrm{len}(\sigma_{\mathrm{home}}) / v_{\mathrm{nom}} \geq T_{\mathrm{thr}} - t$, where $\mathrm{len}(\sigma)$ is the length of the path $\sigma$, $v_{\mathrm{nom}}$ the nominal commanded speed, $T_{\mathrm{thr}}$ the robot's endurance (or mission time limit), $t$ the current time, then the robot is commanded to execute the homing path $\sigma_{\mathrm{home}}$.

            To handle potential unseen obstacles or inadmissible segments of the global path $\sigma_G$, the Local Planning Module is used to track $\sigma_G$. A point $\mathbf{p}_g$ at a distance $d_g$ from the current robot location $\xibf_0$ along $\sigma_G$ is selected as the goal point. The local graph $\Gbb_L$ is built, and the set of shortest paths $\Sigma_L$ from $\xibf_0$ is calculated. The path $\sigma_L \in \Sigma_L$ that takes the robot closest to $\mathbf{p}_g$ is selected and commanded to the robot. Upon execution, $\mathbf{p}_g$ is updated and the process is repeated until the robot reaches the end of $\sigma_G$.

    \subsection{Embodiment Adaptation Layer}\label{subsec:adapt}
        The embodiment adaptation layer specializes the domain-agnostic planning kernel for different robotic platforms by instantiating primitives and criteria for vertex sampling, collision checking, and edge validation during graph construction, parameterized by the robot morphology $R_{\mu}$, motion constraints $C_{\mu}$, and sensing-limited observability constraints. This design preserves a common planning architecture while enabling consistent operation across heterogeneous robotic systems and allowing the same kernel to be reused across multiple embodiments. In its current form, the framework targets platforms whose motion can be reasonably approximated as holonomic or near-holonomic at the planning level. This includes multirotor aerial vehicles, legged and differential-drive ground robots, and hover-capable thruster-actuated underwater robots, such as ROVs or AUVs capable of station keeping and low-speed omnidirectional maneuvering. The supported embodiments are realized through lightweight adaptation mechanisms that modify only the graph-construction primitives while preserving the underlying planning logic.

        \subsubsection{\textbf{Aerial Robot Adaptation}}
            For aerial robots such as multirotors and other rotorcrafts, graph construction is performed directly in the three-dimensional free space encoded by the volumetric map $\mathcal{M}$. In this embodiment, $C_{\mu}$ does not introduce additional constraints beyond those implied by the platform dynamics at the planning-kernel level. Vertices are sampled from the local planning domain as collision-free configurations $\xibf \in \Xi$. A configuration $\xibf$ is accepted if the robot's bounding box $\Bbf_R$ is fully contained within the free space $\Ms_{\mathrm{free}} \subset \Ms$. Given two vertices $\nu_i, \nu_j \in \Vs_k, k\in\{ L,G \}$, an edge $e_{ij}$ is admissible if the straight-line path $\boldsymbol{\gamma}_{ij}(s)~=~(1-s)\xibf_i + s\xibf_j$, $s \in [0,1]$, is collision-free when discretized and evaluated against $\mathcal{M}$.

        \subsubsection{\textbf{Ground Robot Adaptation}}
           For ground robots such as legged systems, graph construction leverages the elevation map $\Hs$ (built following~\cite{fankhauser2018probabilistic}) to enforce terrain support and inclination limits in addition to collision avoidance. These requirements define the ground-specific motion constraints $C_{\mu}$. We refer to the elevation-derived portion as $C_H \subset C_{\mu}$, which requires valid elevation at the footprint query locations and enforces a maximum slope $\theta_{\max}$.

            Each candidate sample $\xibf \in \Xi$ is projected onto $\mathcal{H}$ by querying the elevation at the footprint center and at a set of offsets corresponding to the footprint corners. If elevation data is unavailable or invalid at any queried location, the sample is rejected. For valid projections, the configuration height is set to $z = \mathcal{H}(x,y) + h_0$, where $h_0$ is a nominal clearance representing the robot's height. The sample is accepted only if the inclination between the center and corner elevation values does not exceed a maximum allowable slope $\theta_{\max}$. 
            
            Given two vertices $\nu_i, \nu_j \in \Vs_k, k\in\{ L,G \}$, the candidate edge $e_{ij}$ is evaluated by discretizing the straight-line path between them and projecting each intermediate point onto $\mathcal{H}$. The resulting sequence of projected points defines a ground-consistent polyline $\hat{\boldsymbol{\gamma}}_{ij}$. The edge is rejected if any projected point lacks valid elevation data or if the incremental slope between successive points exceeds $\theta_{\max}$. If $\hat{\boldsymbol{\gamma}}_{ij}$ lies in $\Ms_{\mathrm{free}}$ (using the same evaluation as for aerial robots), $e_{ij}$ is considered admissible, otherwise it is rejected.

        \subsubsection{\textbf{Underwater Robot Adaptation}}
            For underwater robots such as thruster-based \acp{rov}, graph construction follows the same kernel mechanism of sampling vertices within a local planning volume and connecting them via feasible edges as that for aerial robots. However, admissibility criteria are adapted to underwater sensing and operational constraints, while $C_{\mu}$ continues to denote motion constraints only.

            A sampled configuration $\xibf \in \Xi$ is accepted only if it lies in $\Ms_{\mathrm{free}}$ and its Euclidean distance to the nearest occupied voxel in $\mathcal{M}_{\mathrm{occ}}$ is below a predefined proximity threshold $d_{\max}$. This constraint restricts graph expansion to regions sufficiently close to observed structure, thereby prohibiting the robot from entering open-water volumes.

            Given two vertices $\nu_i, \nu_j \in \Vs_k, k\in\{ L,G \}$, a candidate edge $e_{ij}$ is evaluated by discretizing the straight-line path between them and performing collision checking against $\mathcal{M}$ using the robot bounding volume $\Bbf_R$, analogous to that for aerial robots.

    \subsection{Behavior Objectives}
        The planning kernel described in the previous section is task-agnostic and operates solely on the configuration space $\Xi$ and the environment representations $\Ms$ and $\Hs$. Task-specific behaviors are realized by instantiating different objective functions, path evaluation criteria, and termination conditions on top of this shared kernel. Unlike most current planning methods, such as~\cite{kulkarni2022autonomous,dharmadhikari2023gvi,cao2021tare,yang2022far}, that present a monolithic architecture for a single behavior, \acs{omniplanner} operates on the common planning kernel which can facilitate multiple behaviors without modifying its underlying planning structure. In this work, three behavior objectives are considered: (i) \ac{ve}, (ii) \ac{vi}, and (iii) \ac{tr}.
   
        \subsubsection{\textbf{Volumetric Exploration (VE) Behavior}}
            \label{subsec:ve_behavior}
            \begin{algorithm}[t]
                \caption{Volumetric Exploration (VE) Behavior}
                \begin{algorithmic}[1]
                    \Require Current configuration $\xibf_0$, robot morphology $R_{\mu}$, motion constraints $C_{\mu}$, local graph $\mathbb{G}_L$, global graph $\mathbb{G}_G$, volumetric map $\mathcal{M}$, depth sensor model $\mathcal{D}$.
                    \Ensure Exploration path $\sigma_E$.
                    \State $\Sigma_L \gets \textbf{GetShortestPaths}(\xibf_0,\mathbb{G}_L)$
                    \State $\mathbf{G}_{E,L} \gets \textbf{ComputeExplorationGain}( \Sigma_L,\mathcal{D},\mathcal{M})$
                    \State $\sigma_L^\star \gets \textbf{GetBestLocalPath}(R_{\mu},C_{\mu},\Sigma_L, \mathbf{G}_{E,L})$            
                    \If{$\texttt{effectiveExploration}$} 
                        \State $\sigma_E \gets \sigma_L^\star$ \Comment{Local Planner}
                    \Else   \Comment{Global Planner}
                        \State $\Sigma_{G,\mathcal{F}} \gets \textbf{GetShortestPaths}(\nu_{0,G}, \mathcal{F}, \mathbb{G}_G)$
                        \State $\Sigma_{G,\mathrm{home}} \gets \textbf{GetShortestPaths}(\mathcal{F}, \nu_{\mathrm{home}}, \mathbb{G}_G)$
                        \State $\mathbf{G}_{E,G} \gets \textbf{ComputeGlobalGain}(\Sigma_{G,\mathcal{F}}, \Sigma_{G,\mathrm{home}}, \mathcal{M})$
                        \State $\sigma_{G}^\star \gets \textbf{GetBestGlobalPath}(\Sigma_{G,\mathcal{F}}, \mathbf{G}_{E,G})$
                        \State $\sigma_E \gets 
                            \texttt{effectiveExploration}~?~\sigma_{G}^\star:\sigma_{\mathrm{home}}$
                    \EndIf
                    \State \Return $\sigma_E$
                \end{algorithmic}
                \label{alg:exploration}
            \end{algorithm}
            
            The \ac{ve} behavior instantiates the planning kernel with the objective of incrementally classifying the environment volume using depth sensing $\mathcal{D}$, under the robot’s motion and sensing constraints.

            \begin{behavior}[Volumetric Exploration] 
                Given a bounded environment volume $V$ and an initial robot configuration $\xibf_{\mathrm{init}} \in \Xi$, determine a collision-free path $\sigma_E$ that enables the classification of the environment into free space $V_{free} \subset V$ and occupied space $V_{occ} \subset V$, based on observations acquired by the depth sensor $\mathcal{D}$. Exploration is considered complete when no further reachable, collision-free configuration exists from which any remaining unclassified portion of the environment can be observed, i.e., $V_{free} \cup V_{occ} = V \setminus V_{\mathrm{res}}$. The generated paths must satisfy the robot’s motion constraints $C_{\mu}$ at all times.  
            \end{behavior}

            Let $\mathbb{G}_L = \{\mathcal{V}_L, \mathcal{E}_L\}$ denote the local planning graph constructed by the planning kernel, rooted at the current robot configuration $\xibf_0$. At each planning iteration, the shortest paths $\Sigma_L$ are calculated using Dijkstra's algorithm from the root vertex $\nu_0$, corresponding to the configuration $\xibf_0$, to all vertices in the graph. For a configuration $\xibf \in \Xi$, let $\Gamma_{\mathrm{VE}}(\xibf)$ denote the volumetric information gain, defined as the number of previously unknown voxels in $\mathcal{M}$ that would become observable by the depth sensor $\mathcal{D}$ if the robot were to be in the configuration $\xibf$, accounting for sensor \acs{fov} $[F_H^{\mathcal{D}}, F_V^{\mathcal{D}}]$, maximum range $d_{\max}^{\mathcal{D}}$, and visibility constraints.

            The exploration objective evaluates candidate paths $\sigma_L\in\Sigma_L$ based on the cumulative information gain along the path:
            \begin{equation}
                \mathcal{J}_{\mathrm{VE}}^L(\sigma_L) = e^{-\mu_d \lambda_d(\sigma_L)} \sum_{k=1}^{N} \Gamma_{\mathrm{VE}}(\xibf_k) e^{-\mu_l \lambda_l(\nu_0, \nu_k)},
            \end{equation}
            where $\lambda_l(\nu_0,\nu_k)$ denotes the path length from the root $\nu_0$ to vertex $\nu_k$ along $\Gbb_L$ and $\mu_l > 0$ is a distance-based penalty factor. The direction penalty $\lambda_d(\sigma_L)$, with the direction-based penalty factor $\mu_d>0$, penalizes paths that deviate from the current exploration direction. It is defined as the normalized \acs{dtw} distance between the candidate path $\sigma_L$ and a straight reference path $\sigma_{\exp}$ aligned with the exploration direction, i.e., $\lambda_d(\sigma_L) = \delta(\sigma_L, \sigma_{\exp})$. Among all feasible paths $\sigma_L \in \Sigma_L$ extracted from the local planning graph, the \ac{ve} behavior selects
            \begin{equation}
                \sigma_L^\star = \arg\max_{\sigma_L\in\Sigma_L}\mathcal{J}_{\mathrm{VE}}^L(\sigma_L).
            \end{equation}
            
            The \ac{ve} behavior terminates locally when all candidate paths in the local planning graph $\mathbb{G}_L$ yield negligible cumulative information gain, i.e., when $\sum_{k=1}^{N}\Gamma_{\mathrm{VE}}(\xibf_k)$ is negligible for all $\sigma_L\in\Sigma_L$, indicating that no further reduction of environmental uncertainty is achievable within the local planning volume (\texttt{effectiveExploration~=~False} in line 6 of Alg.~\ref{alg:exploration}). In this case, the global planning graph $\mathbb{G}_G$ is queried to compute a collision-free repositioning path to another previously explored region. This maneuver is called ``Global Repositioning''.
            In the \ac{ve} behavior, the global graph maintains a set of vertices $\Fs$ called ``frontier'' vertices that have $\Gamma_{\mathrm{VE}} > \Gamma_{\mathrm{thr}, \Fs}$, where $\Gamma_{\mathrm{thr}, \Fs}$ is the threshold on the volumetric information gain for a vertex to qualify as frontier. When the local exploration is exhausted, the planner repositions the robot to one of the vertices in $\Fs$. To select the best frontier, first, the set $\Sigma_{G,\Fs}$ of the shortest paths from the vertex $\nu_{0,G} \in \Vs_G$ corresponding to $\xibf_0$ to all vertices in $\Fs$ is calculated. Next, the shortest paths $\Sigma_{G,\mathrm{home}}$ from each vertex in $\Fs$ to the home vertex $\nu_{\mathrm{home}}$ are calculated. To select the best frontier vertex, a Global Gain is calculated for each vertex $\nu_f \in \Fs$ as:
            \begin{equation}
                \mathcal{J}_{\mathrm{VE}}^G(\nu_f) = \Ts(\nu_{0,G}, \nu_f) \Gamma_{\mathrm{VE}}(\xibf_f) e^{-\mu_l \lambda_l(\nu_{0,G}, \nu_f)},
            \end{equation}
            where $\xibf_f$ is the robot configuration corresponding to $\nu_f$, $\mu_l$ and $\lambda_l$ are same as those for $\mathcal{J}_{\mathrm{VE}}^L$, and $\Gamma_{\mathrm{VE}}$ is the volumetric information gain. $\Ts(\nu_{0,G}, \nu_f)$ is a function that estimates the remaining exploration time and is defined as:
            \begin{equation}
                \Ts(\nu_{0,G}, \nu_f) = T_{\mathrm{thr}} - \frac{\lambda_l(\nu_{0,G}, \nu_f)}{v_{\mathrm{nom}}} - \frac{\lambda_l(\nu_f, \nu_{\mathrm{home}})}{v_{\mathrm{nom}}}.
            \end{equation}
            If no frontiers exist in $\Gbb_G$, the planner concludes that the exploration is complete, calculates a safe path $\sigma_{\mathrm{home}}$ from $\nu_{0,G}$ to $\nu_{\mathrm{home}}$, and commands to the robot to execute it.

        \subsubsection{\textbf{Visual Inspection (VI) Behavior}}
            \label{subsec:vi_behavior}
            \begin{algorithm}[t]
                \caption{Visual Inspection (VI) Behavior}
                \begin{algorithmic}[1]
                    \Require Current configuration $\xibf_0$, robot morphology $R_{\mu}$, motion constraints $C_{\mu}$, local graph $\mathbb{G}_L$, volumetric map $\mathcal{M}$, camera sensor model $\mathcal{C}$.
                    \Ensure Inspection path $\sigma_I$.
                    \State $P_I \gets \textbf{SamplePointsInVolume}(\Bbf_{\mathrm{VI}},\mathcal{M})$
                    \State $\Xi_I \gets \emptyset$
                    \ForAll{$\pbf_v \in P_I$}
                        \If{$d_{\min}^{\mathcal{C}} \le \textbf{SDF}(\pbf_v) \le d_{\max}^{\mathcal{C}}$}
                            \State $O_v \gets \textbf{ComputeOrientations}(\pbf_v,\Ms_I,\mathcal{C})$
                            \ForAll{$\langle \psi,\vartheta_a\rangle \in O_v$}
                                \State $\xibf_v \gets [p_v^x,p_v^y,p_v^z,\psi,\vartheta_a]$
                                \State $\Xi_I \gets \Xi_I \cup \{\xibf_v\}$
                            \EndFor
                        \EndIf
                    \EndFor
                    \State $\Xi_I^\star \gets \textbf{GreedyCoverageSelection}(\Xi_I, S_I, \mathcal{C}, \mathcal{M})$
                    \If{$\Xi_I^\star = \emptyset$}
                        \State \Return $\emptyset$
                    \EndIf
                    
                    \State $\Gbb_L \gets \textbf{BuildLocalGraph}(\xibf_0,R_{\mu},C_{\mu},\mathcal{M},\Bbf_L=\Bbf_{\mathrm{VI}})$
                    \State $\Vs_I^\star \gets \textbf{InsertAndConnectViewpoints}(\Gbb_L,\Xi_I^\star)$
                    \State $\mathbf{D} \gets \textbf{AllPairsShortestPathLengths}(\Gbb_L,\Vs_I^\star)$
                    \State $\pi^\star \gets \textbf{SolveTSP}(\mathbf{D})$ \Comment{LKH~\cite{helsgaun2000effective}}
                    \State $\sigma_I^\star \gets \textbf{ConcatenateShortestPaths}(\Gbb_L,\pi^\star,\Vs_I^\star)$
                     \State $\sigma_I \gets \sigma_I^\star$
                    \State \Return $\sigma_I$
                \end{algorithmic}
                \label{alg:inspection}
            \end{algorithm}

           The \ac{vi} behavior addresses the problem of systematically observing a specified subset of visible surface regions in the environment while respecting camera sensing constraints. The inspection task is formally defined as follows.

            \begin{behavior}[Visual Inspection]  
            	Given a target surface set $S_I$ related to the associated target volumetric map $\Ms_I \subset \mathcal{M}_{\mathrm{occ}}$ to be inspected, determine a collision-free path $\sigma_I$ such that the camera sensor $\mathcal{C}$ observes all elements of $S_I$ within its \ac{fov} $[F^{\Cs}_H, F^{\Cs}_V]$ and effective sensing range $[d_{\min}^{\mathcal{C}}, d_{\max}^{\mathcal{C}}]$. The inspection process terminates when no collision-free configuration exists from which any remaining unobserved surface region $S_I \setminus S_{\mathrm{res}}$ can be perceived.  
            \end{behavior}
            
            Candidate inspection viewpoints are generated in the free space surrounding $S_I$ using $\Ms$. First, a set $P_I$ of $3$D points is sampled within a bounded inspection domain $\Bbf_{\mathrm{VI}}$ enclosing $\mathcal{M}_I$. 
            A point $\pbf_v \in P_I$ is accepted if its signed distance satisfies $d_{\min}^{\mathcal{C}} \leq \text{SDF}(\pbf_v) \leq d_{\max}^{\mathcal{C}}$, ensuring collision-free placement within the effective sensing range. 
            For each point $\pbf_v = [p_v^x, p_v^y, p_v^z]$, the set $O_v$ of robot orientation and camera pitch (if available) combinations $\{\psi, \vartheta_a\}$ is computed such that all occupied voxels $m \in \Ms_I$ lying within a solid spherical shell centered at $\pbf_v$, with inner and outer radii $d^C_{\min}$ and $d^C_{\max}$ respectively, are observable.
            A set $\Xi_v$ of viewpoint configurations $\xibf_v^i$ corresponding to each $\langle \psi_v^i, \vartheta_{a,v}^i \rangle ~\in~ O_v$ is generated such that $\xibf_v^i = [p_v^x, p_v^y, p_v^z, \psi_v^i, \vartheta_{a,v}^i]$. The set $\Xi_I$ of all viewpoint candidates for the \ac{vi} behavior is:
            \begin{equation}
                \Xi_I = \bigcup_{\pbf_v \in P_I} \Xi_v.
            \end{equation}
            To reduce redundancy, the initial viewpoint set $\Xi_I$ is reduced to a coverage-optimal subset $\Xi_I^\star$ using a greedy gain-driven strategy. The objective of this reduction is to iteratively select viewpoints that maximize the marginal coverage of previously unobserved surface elements. Starting from $\Xi_I^0 = \emptyset$, at each iteration $k$, the viewpoint with the largest marginal gain is selected as:
            \begin{equation}
                \xibf_k^\star =
                \arg\max_{\xibf \in \Xi_I \setminus \Xi_I^{k-1}}
                \left|
                \mathrm{Vis}(\xibf) \setminus
                \bigcup_{\xibf' \in \Xi_I^{k-1}} \mathrm{Vis}(\xibf')
                \right|,
            \end{equation}
            where $\mathrm{Vis}(\xibf)$ denotes the subset of $S_I$ visible from configuration $\xibf$ under the sensor constraints. The set is updated as $\Xi_I^k = \Xi_I^{k-1} \cup \{\xibf_k^\star\}$. The process terminates when the incremental gain falls below a threshold $\epsilon_I$, or when all observable surface elements have been covered. The final reduced viewpoint set is then given by $\Xi_I^\star = \Xi_I^k$. Importantly, this procedure does not require selected viewpoints to have mutually disjoint FoVs. Overlapping surface elements are treated as redundant and are excluded from the marginal-gain computation. Therefore, each candidate viewpoint is scored only by the additional, previously unobserved surface elements it covers.
            
            To enable path planning between selected viewpoints, a local planning graph $\Gbb_L$ is constructed by sampling collision-free configurations within $\Bbf_{\mathrm{VI}}$ using the Local Planning Module with $\Bbf_L = \Bbf_{\mathrm{VI}}$. For each viewpoint $\xibf_v \in \Xi_I^\star$, a vertex $\nu_v$ is explicitly inserted and connected to nearby vertices of $\Gbb_L$ via admissible edges forming the set of viewpoint vertices $\Vs_I^\star$.

            The inspection trajectory is obtained by solving a shortest path problem over $\Gbb_L$ that visits all viewpoints in $\Xi_I^\star$. The optimal inspection path
            \begin{equation}
                \sigma_I^\star = \arg\min_{\sigma \in \Sigma_I} \sum_{k=1}^{|\sigma|-1} d_l(\xibf_k, \xibf_{k+1}),
            \end{equation}
            where $|\sigma|$ is the number of configurations in $\sigma$ and $d_l(\xibf_k,\xibf_{k+1})$ denotes the path length between successive configurations. The solution is required to satisfy the coverage constraint
            \begin{equation}
                \bigcup_{\xibf_k \in \sigma} \mathrm{Vis}(\xibf_k) \supseteq \mathcal{M}_{\mathrm{occ}} \setminus S_{\mathrm{res}},
            \end{equation}
            where $\mathrm{Vis}(\xibf_k)$ denotes the surface region visible from $\xibf_k$ and $S_{\mathrm{res}}$ is the residual surface defined in Section~\ref{sec:problem_statement}. The planner calculates $\sigma_I^{\star}$ by solving the \ac{tsp} to find the ordering between the viewpoints $\xibf_v \in \Xi_I^{\star}$. The cost of traveling $d(\xibf_i, \xibf_j)$ between $\xibf_i, \xibf_j \in \Xi_I^{\star}$ is the length of the shortest path between the corresponding vertices $\nu_i, \nu_j \in \Vs_I^{\star}$ in $\Gbb_L$. We utilize the Lin-Kernighan-Helsgaun (LKH) heuristic~\cite{helsgaun2000effective} to solve the \ac{tsp}.
            The final inspection trajectory is obtained by concatenating the shortest collision-free paths between successive viewpoints along $\sigma_I^\star$. The inspection behavior terminates when no additional collision-free viewpoints yield positive visual gain, indicating that all observable surface regions have been inspected.

        \subsubsection{\textbf{Target Reach (TR) Behavior}}
            \label{subsec:tr_behavior}
            \begin{algorithm}[t]
                \caption{Target Reach (TR) Behavior}
                \begin{algorithmic}[1]    
                    \Require Current configuration $\xibf_0$, robot morphology $R_{\mu}$, motion constraints $C_{\mu}$, local graph $\mathbb{G}_L$, volumetric map $\mathcal{M}$.
                    \Ensure Target path $\sigma_T$.
                    \If{$\|\mathrm{pos}(\xibf_0)-\pbf_t\|_2 \le \rho_{\mathrm{reach}}$}
                        \State \Return $\emptyset$ \Comment{Target reached}
                    \EndIf
                    
                    \State $\nu_0 \gets \textbf{GetGlobalVertex}(\Gbb_G,\xibf_0)$
                    \State $\nu_c \gets \textbf{NearestVertexWithinRadius}(\Gbb_G,\pbf_t,\rho_t)$
                    \If{$\nu_c \neq \emptyset$}
                        \State $\nu_{\mathrm{best}} \gets \nu_c$
                    \Else
                        \If{$\Fs = \emptyset$}
                            \State \Return $\emptyset$ \Comment{No frontier to progress}
                        \EndIf
                        \State $\nu_{\mathrm{best}} \gets \textbf{GetBestVertex}(\pbf_t,\nu_0,\Fs,\Gbb_G,\mathcal{M})$
                    \EndIf
                    \State $\sigma_{\mathrm{guide}} \gets \textbf{GetShortestPath}(\Gbb_G,\nu_0,\nu_{\mathrm{best}})$
                    \State $\pbf_{\mathrm{lh}} \gets \textbf{SelectLookaheadPoint}(\sigma_{\mathrm{guide}},\rho_{\mathrm{lh}})$
                    \If{$\pbf_t \in \Bbf_L$}
                        \State $\pbf_{\mathrm{lh}} \gets \pbf_t$
                    \EndIf
                    \State $\Sigma_L \gets \textbf{GetShortestPaths}(\Gbb_L,\nu_0)$
                    \State $\sigma_T^\star \gets \textbf{GetBestPathToLookahead}(R_{\mu},C_{\mu},\Sigma_L,\pbf_{\mathrm{lh}})$
                    \State $\sigma_T \gets \sigma_T^\star$
                    
                    \State \Return $\sigma_T$
                \end{algorithmic}
                \label{alg:target_reach}
            \end{algorithm}

            The \ac{tr} behavior instantiates the planning kernel with the objective of guiding the robot toward a user-defined target position $\pbf_t \in \mathbb{R}^3$ (potentially in unknown space), while respecting the robot’s motion and sensing constraints. Unlike conventional methods \cite{ren2025safety}, \acs{omniplanner} is able to reach targets in unknown space in complex, $3$D environments (e.g., Fig.~\ref{fig:aerial_robot_target_reach}) due to the bifurcated local-global architecture of the planning kernel.

            \begin{behavior}[Target Reach]  
            	Given a bounded environment volume $V$, an initial robot configuration $\xibf_{\mathrm{init}} \in \Xi$, and a predefined target position $\pbf_t \in \mathbb{R}^3$, determine a collision-free path $\sigma_T$ that guides the robot toward the target. The selected path must satisfy the robot’s motion constraints $C_{\mu}$ at all times and is chosen such that its terminal configuration minimizes the Euclidean distance to the target. The target reach behavior is considered complete when the robot reaches the target within a predefined tolerance $\rho_{\mathrm{reach}}$ or when no further reachable, collision-free configuration exists that reduces the distance to the target.  
            \end{behavior}

            In the \ac{tr} behavior, the planner iteratively calculates paths to take the robot closer to the target $\pbf_t$. Let $\Gbb_G = \{ \Vs_G, \Es_G \}$ be the graph built by the Global Planning Module, and $\Gbb_L = \{ \Vs_L, \Es_L \}$ be the graph built by the Local Planning Module of the planning kernel in each planning iteration. Similar to the \ac{ve} behavior, the planner keeps track of the set $\Fs$ of frontier vertices in $\Gbb_G$ that have the volumetric gain $\Gamma_{\mathrm{VE}}~>~\Gamma_{\mathrm{thr},\Fs}$. In each planning iteration, the planner selects the best vertex $\nu_{\mathrm{best}}$ in $\mathbb{G}_G$ to advance toward $\mathbf{p}_t$, calculates a guiding path $\sigma_{\mathrm{guide}}$ towards it, and then computes a collision-free path, using the Local Planning Module, to advance along the guiding path. 
            
            More specifically, the planner first checks if $\pbf_t$ is close to the explored space by searching for the closest vertex $\nu_c$ to $\pbf_t$ within a distance $\rho_t$ in $\Gbb_G$. If found, $\nu_{\mathrm{best}} = \nu_c$ and the shortest path from the vertex $\nu_0$, corresponding to the current robot configuration $\xibf_0$, to $\nu_{\mathrm{best}}$ along $\Gbb_G$ is calculated and used as the guiding path $\sigma_{\mathrm{guide}}$ to reach $\pbf_t$.
            If $\nu_c$ is not found, then $\pbf_t$ is sufficiently far away from the explored space. In this case, the planner finds the best frontier vertex in $\Gbb_G$ to visit to make progress towards the target. For a frontier vertex $\nu_f$, let $d_f^l$ be the shortest path length from $\nu_0$ to $\nu_f$ along $\Gbb_G$ and $d_f^u$ be the Euclidean distance between the position of $\nu_f$ and $\pbf_t$. Then, the best frontier to visit is selected as follows:
            \begin{equation}
                \nu_{\mathrm{best}} = \arg\min_{\nu_f \in \mathcal{F}} \; d_f^l + \lambda_{\mathrm{bal}} d_f^u,
            \end{equation}
            where $\lambda_{\mathrm{bal}} = 1 + \frac{d_f^u}{d_f^u + d_f^l}$ is a balancing term that trades off the cost of reaching a frontier against its proximity to $\pbf_t$. The shortest path from the vertex $\nu_0$ to $\nu_{\mathrm{best}}$ along $\Gbb_G$ is calculated and used as the guiding path $\sigma_{\mathrm{guide}}$ to reach $\pbf_t$.

            Once the guiding path is calculated, the Local Planning Module is used to calculate a local, collision-free path to reach $\nu_{\mathrm{best}}$. First, a lookahead point $\pbf_{\mathrm{lh}}$ on $\sigma_{\mathrm{guide}}$ is selected at a distance $\rho_{\mathrm{lh}}$ along the path (note that this is the distance along the path, not Euclidean distance between the current robot location and the point on the path). If the target position $\pbf_t$ is within the local planning volume $\Bbf_L$, then $\pbf_{\mathrm{lh}} = \pbf_t$.
            Dijkstra’s algorithm is applied to the local planning graph $\mathbb{G}_L$ to compute the shortest paths $\Sigma_L$ from the root configuration $\xibf_0$ to all vertices in the graph. For a configuration $\xibf \in \Xi$, let $\Gamma_{\mathrm{TR}}(\xibf)$ denote the lookahead distance metric, defined as the Euclidean distance between the position component of $\xibf$ and the lookahead position $\pbf_{\mathrm{lh}}$:  
            \begin{equation}
                \Gamma_{\mathrm{TR}}(\xibf) = \left| \mathrm{pos}(\xibf) - \pbf_{\mathrm{lh}} \right|_2.
            \end{equation}
            The \ac{tr} objective evaluates candidate paths $\sigma_L \in \Sigma_L$ based on the distance between the lookahead and the terminal (leaf) configuration $\xibf_N$ of the path:  
            \begin{equation}
                \mathcal{J}_{\mathrm{TR}}(\sigma_L) = \Gamma_{\mathrm{TR}}(\xibf_N),
            \end{equation}
            where $\xibf_N$ denotes the final configuration along the path $\sigma_L$. Among all feasible paths $\sigma_L \in \Sigma_L$ extracted from $\Gbb_L$, the \ac{tr} behavior selects  
            \begin{equation}  
                \sigma_T^\star = \arg\min_{\sigma_L \in \Sigma_L} \mathcal{J}_{\mathrm{TR}}(\sigma_L).
            \end{equation}
            The \ac{tr} behavior terminates when either a) the robot reaches $\pbf_t$ within a user-defined distance $\rho_{\mathrm{reach}}$, b) no frontier exists ($\Fs = \emptyset$), or c) no terminal vertex in $\Sigma_L$ takes the robot closer to $\pbf_t$ for $n$ consecutive planning iterations.    
    
     \section{SIMULATION STUDIES}
    \label{sec:simulation_studies}
    This section presents simulation-based validation of the proposed planning framework. We first perform a feature-specific evaluation to qualitatively assess the impact of key design choices in the planning kernel. We then demonstrate the behavior of the complete system in representative simulation scenarios. Finally, we quantitatively compare the proposed framework against state-of-the-art planning methods.

    \subsection{Planning Feature Evaluation}
        This subsection provides evaluations of individual planning features. The objective of this study is to demonstrate how specific design choices influence the planner's behavior and performance under representative conditions. 
        \begin{figure*}[ht!]
            \centering
            \includegraphics[clip, trim = 0cm 0cm 0cm 0cm, width=1\linewidth]{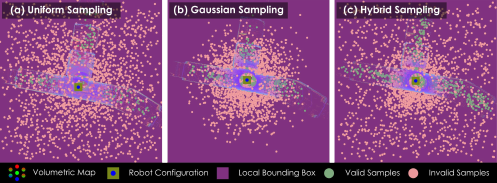}
            \caption{Indicative visualization of local sampling strategies within the bounded planning volume. Green points denote collision-free valid samples, yellow points indicate invalid samples rejected during collision checking, and the underlying point cloud represents the current volumetric map. (a) Uniform sampling distributes configurations evenly throughout the local bounding box, promoting broad spatial coverage. (b) Gaussian sampling concentrates samples around the current robot configuration, yielding dense local connectivity in constrained regions. (c) Hybrid sampling combines Uniform and Gaussian distributions to balance local maneuverability and global reach.}
            \label{fig:sampling_strategies}
        \end{figure*}
        \subsubsection{\textbf{Sampling Strategies}}
            The Local Planning Module supports three sampling strategies within the bounded planning volume $\mathbf{B}_L$: (i) Uniform, (ii) Gaussian, and (iii) Hybrid. All strategies operate within the same $\mathbf{B}_L$ and differ only in how candidate configurations are distributed before collision checking and graph construction. These distributions introduce different spatial biases, affecting the local density and outward reach of the resulting graph. To illustrate this behavior, we consider the representative T-shaped corridor environment in Fig.~\ref{fig:sampling_strategies}, which contains both narrow passages and branching connectivity.

            Fig.~\ref{fig:sampling_strategies} shows one representative sampling instance for each strategy, while Table~\ref{tab:sampling_diagnostic} provides a quantitative diagnostic from a fixed robot configuration inside a $20 \times 20 \times 10$ m local planning volume. For each strategy, $100$ candidate samples are generated per trial over $20$ randomized trials. The table reports the mean candidate distance from the robot, the number of local samples within $5$ m, and the number of exploratory samples beyond $5$ m.

            Uniform sampling spreads candidates broadly across $\mathbf{B}_L$, promoting outward reach toward distant branches and open regions, but it can be inefficient in constrained geometries because many samples may be rejected during collision checking. Gaussian sampling concentrates candidates around the current robot configuration $\xibf_0$, improving local sample density in narrow regions, but reducing coverage of distant parts of the planning volume. Hybrid sampling combines these effects by placing part of the candidates near the robot while maintaining broader coverage.

        \subsubsection{\textbf{Graph Construction Strategies}}
            This subsection evaluates the two local graph construction strategies introduced in Section~\ref{sec:planning_kernel}, namely the Incremental construction and the Batch construction methods. Both strategies operate on the same bounded planning volume $\mathbf{B}_L$, and differ only in how the collision-free vertices are added and connected during graph generation. To study their effect on reachable-space representation, we consider two representative environments: (i) a multi-room building layout consisting of six interconnected rooms arranged around a central corridor, and (ii) a multi-branch mine topology containing six tunnel branches connected at a junction. For each strategy, experiments are performed with eight different sample counts ranging from $100$ to $800$, each repeated over 20 trials. For every trial, we record the computation time required to build the local graph and the number of distinct reachable regions discovered by the graph (rooms or branches). 
            \begin{table}[t!]
                \centering
                \caption{Sampling distribution diagnostic in a local bounding box ($20~\times~20~\times~10~\mathrm{m}$) over $20$ sampling trials.}
                \label{tab:sampling_diagnostic}
                \setlength{\tabcolsep}{6.0pt}
                \renewcommand{\arraystretch}{1.3}
                \begin{tabular}{lccc}
                    \toprule
                    \textbf{\shortstack{Sampling\\Method}} &
                    \textbf{\shortstack{Mean\\distance (m)}} &
                    \textbf{\shortstack{Local\\samples ($\#$)}} &
                    \textbf{\shortstack{Exploratory\\samples ($\#$)}} \\
                    \hline
                    Uniform &
                    $8.31 \pm 0.27$ &
                    $13.65 \pm 3.41$ &
                    $86.35 \pm 3.41$ \\
                    Gaussian, $\sigma=3$ &
                    $4.52 \pm 0.21$ &
                    $62.10 \pm 5.26$ &
                    $37.90 \pm 5.26$ \\
                    Hybrid &
                    $6.41 \pm 0.23$ &
                    $39.25 \pm 3.08$ &
                    $60.75 \pm 3.08$ \\
                    \bottomrule
                \end{tabular}
            \end{table}
            \begin{figure}[t!]
                \centering
                \includegraphics[clip, trim = 0cm 0cm 0cm 0cm, width=1\linewidth]{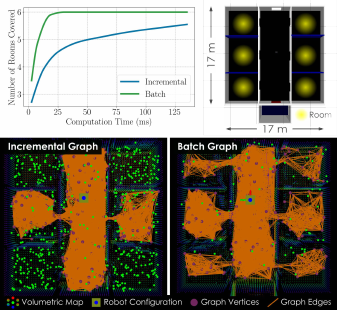}
                \caption{Comparison of local graph construction strategies in a multi-room building environment. Top-left: average number of rooms covered as a function of computation time for Incremental and Batch graph construction. For each strategy, experiments were performed with eight different numbers of sampled points, each repeated over 20 trials. The plotted curves report the mean computation time and the mean number of reachable rooms covered by the local graph. Top-right: layout and dimensions of the evaluation environment. Bottom: representative local graph instances generated by the two strategies within the same planning volume.}
                \label{fig:graph_construction_rooms}
            \end{figure}
            \begin{figure}[t!]
                \centering
                \includegraphics[clip, trim = 0cm 0cm 0cm 0cm, width=1\linewidth]{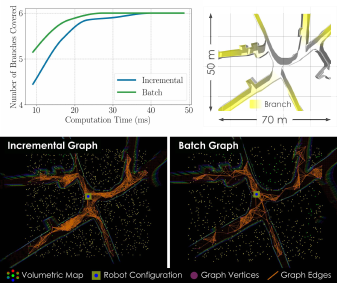}
                \caption{Comparison of local graph construction strategies in a multi-branch mine environment. Top-left: average number of branches covered as a function of computation time for Incremental and Batch graph construction. For each strategy, experiments were performed with eight different numbers of sampled points, each repeated over 20 trials. The plotted curves report the mean computation time and the mean number of reachable branches covered by the local graph. Top-right: layout and dimensions of the evaluation environment. Bottom: representative local graph instances generated by the two strategies within the same planning volume.}
                \label{fig:graph_construction_mine}
            \end{figure}
            
            The curves in Fig.~\ref{fig:graph_construction_rooms} and Fig.~\ref{fig:graph_construction_mine} report the computation time and the mean number of reachable regions covered by each strategy. The Batch method identifies multiple reachable regions more quickly, particularly at lower computation times, owing to its broader sampling of the planning volume prior to edge pruning. In contrast, the Incremental strategy expands incrementally from the current robot configuration, requiring more time to reach distant regions but producing structured graph growth that closely follows feasible corridors. As the number of samples increases, both strategies converge to similar coverage once all reachable regions are discovered. The bottom visualizations in Figs.~\ref{fig:graph_construction_rooms} and~\ref{fig:graph_construction_mine} show representative local graph instances generated by the two strategies within the same planning volume. These examples illustrate the qualitative difference between the methods: Incremental construction produces corridor-following branches that reflect reachable paths, while Batch construction yields denser connectivity and faster region discovery in both multi-room and multi-branch environments.

            To complement the graph-level analysis, we further evaluate how the two graph-construction strategies affect mission-level exploration performance. We compare four variants: Incremental, Batch, Incremental with planning-while-flying (Incremental + PWF), and Batch with planning-while-flying (Batch + PWF). All variants use the same robot model, sensing model, local planning volume, sampling budget, graph-size limits, exploration objective, and environment. Therefore, differences between Incremental and Batch under the same execution mode reflect the effect of graph construction in the considered environment, while differences between the stop-plan-go and PWF variants reflect the effect of overlapping planning with motion execution.

\begin{figure}[t!]
                \centering
                \includegraphics[clip, trim = 0.0cm 0cm 0cm 0cm, width=1\linewidth]{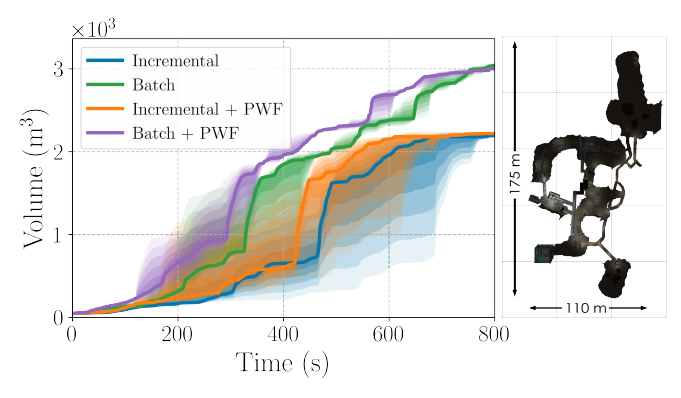}
                \caption{Ablation of graph construction and planning-while-flying (PWF) during aerial exploration. Incremental and Batch evaluate the effect of the local graph-construction strategy under stop-plan-go execution, while Incremental + PWF and Batch + PWF use the same graph-construction strategies with planning performed concurrently during motion execution. Solid curves denote median explored volume over time, and shaded regions indicate variation across trials.}
                \label{fig:planning_pwf_ablation}
            \end{figure}

            Fig.~\ref{fig:planning_pwf_ablation} reports the explored volume over time for the four variants over three trials. Batch graph construction improves exploration progress compared with Incremental construction under the same execution mode, indicating that the graph-construction strategy affects how quickly useful reachable regions are discovered. PWF further improves both graph-construction variants by reducing idle time between planning cycles and path execution. The combined Batch + PWF variant achieves the fastest coverage growth in this experiment, while Incremental stop-plan-go shows the slowest progress. The magnitude of these differences is environment-dependent. Batch construction is most beneficial when the local planning volume contains branching connectivity, multiple reachable regions, or free-space pockets that can be discovered through broader connectivity testing.

        \subsubsection{\textbf{Camera Sensing Strategies}}
            \begin{figure*}[t]
                \centering
                \includegraphics[clip, trim = 0cm 0cm 0cm 0cm, width=1\linewidth]{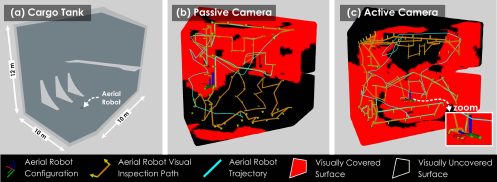}
                \caption{Evaluation of passive and active camera configurations for aerial inspection in a cargo tank environment. (a) Cargo tank model used in the simulation study. (b) Inspection using a passive camera configuration. (c) Inspection using an active camera configuration with an actuated camera supporting upward- and downward-looking viewpoints. The illustrated instances correspond to representative stages of mission execution and demonstrate the progressive accumulation of inspected surface coverage. The passive camera configuration achieves a coverage of $62.16 \pm 9.01\%$, whereas the active camera configuration increases the coverage to $88.01 \pm 5.36\%$.}
                \label{fig:camera_strategies}
            \end{figure*}

            To assess the impact of camera actuation on the \ac{vi} behavior, we compare two sensing strategies in a cargo tank inspection scenario (Fig.~\ref{fig:camera_strategies}a): (i) a Passive (body-fixed) Camera and (ii) an Active Camera with actuated pitch. Both strategies follow the inspection pipeline described in Section~\ref{subsec:vi_behavior} and differ only in the camera model used to instantiate $\mathrm{Vis}(\cdot)$ and the associated viewpoint orientation set $O_v$. With a passive camera, the sensor is rigidly mounted with a fixed pitch. Consequently, each sampled position $\pbf_v$ admits an orientation set $O_v$ that varies only in yaw. This limits the visible surface per viewpoint and typically requires additional viewpoints to mitigate occlusions and unfavorable viewing angles (Fig.~\ref{fig:camera_strategies}b). In contrast, the active camera allows pitch actuation, and $O_v$ includes feasible yaw--pitch pairs $\{\psi,\vartheta_a\}$ that satisfy the camera \ac{fov} and range constraints. This expands the achievable visibility from the same spatial samples, allowing the greedy selection to cover larger portions of $S_I$ with fewer redundant viewpoints and yielding more effective inspection routes (Fig.~\ref{fig:camera_strategies}c).
        
            To evaluate the inspection benefits of the active camera configuration, we conducted a simulation study in a cargo tank measuring $10 \times 10 \times 12$ m, as illustrated in Fig.~\ref{fig:camera_strategies}a. Five independent trials were performed for each strategy using identical planning budgets and inspection parameters. Coverage is reported as the percentage of the target surface observed along the entire executed inspection trajectory. The passive camera configuration achieves a coverage of $62.16 \pm 9.01\%$, whereas the active camera increases coverage to $88.01 \pm 5.36\%$. Fig.~\ref{fig:camera_strategies}b-c show representative instances acquired during the execution of the inspection mission. The cyan trajectory denotes the robot path executed up to the illustrated time instance, while the orange path represents the inspection trajectory generated by the planner. These instances provide a qualitative illustration of how surface coverage accumulates throughout the mission and highlight the improved inspection performance achieved by the active camera configuration, which employs an actuated camera capable of both upward- and downward-facing observations.

    \subsection{Simulation Scenarios}
        We conduct a set of evaluations on aerial, ground, and underwater robots demonstrating the performance of \acs{omniplanner} across morphologies, environments and tasks.

        \subsubsection{\textbf{Aerial Robot: Target Reach Behavior}}
            \begin{figure*}[t]
                \centering
                \includegraphics[clip, trim = 0cm 0cm 0cm 0cm, width=1\linewidth]{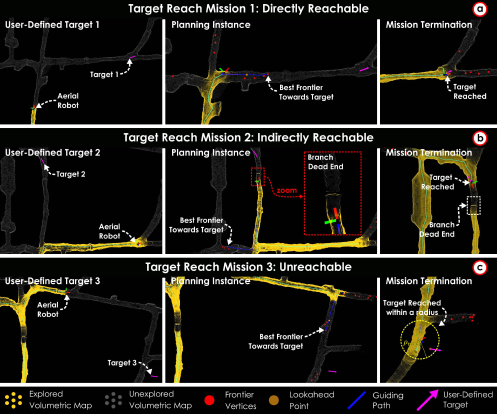}
                \caption{Target reach behavior for an aerial robot in a cave environment under three representative missions. Each row shows (left) the user-defined target location in the partially explored map, (middle) a planning instance highlighting the selected best frontier (red) and the global guiding path $\sigma_{\mathrm{guide}}$ (blue) with its lookahead point $\pbf_{\mathrm{lh}}$ (orange), and (right) the mission termination state. (a) Directly reachable target. (b) Indirectly reachable target requiring global frontier selection to avoid a dead-end branch. (c) Unreachable target: the planner advances toward the closest attainable region and terminates when no further progress is possible, within the tolerance radius $\rho_{\mathrm{reach}}$.}
                \label{fig:aerial_robot_target_reach}
            \end{figure*}

            We demonstrate the proposed \ac{tr} behavior in a cave-like environment using three representative missions (Fig.~\ref{fig:aerial_robot_target_reach}). In all cases, a user-defined target position $\pbf_t$ is specified in initially unknown space. The planner first reasons on the global graph $\Gbb_G$ to select a guiding point, either a vertex near $\pbf_t$ when it becomes reachable from the explored space, or an intermediate frontier vertex that best advances toward $\pbf_t$ when the target is still far from any explored region. A guiding path $\sigma_{\mathrm{guide}}$ is then computed on $\Gbb_G$, and the Local Planning Module generates collision-free motions by tracking a lookahead point $\pbf_{\mathrm{lh}}$ along $\sigma_{\mathrm{guide}}$.

            Fig.~\ref{fig:aerial_robot_target_reach}a illustrates Mission~1, where the user-defined target lies outside the currently explored volume but is directly reachable in the sense that progress toward $\pbf_t$ does not require intermediate global repositioning (e.g., detours to alternative branches). The planner expands exploration along the guiding direction until the target becomes reachable and terminates once the robot arrives within the user-defined tolerance $\rho_{\mathrm{reach}}$. Fig.~\ref{fig:aerial_robot_target_reach}b shows Mission~2, where the target is indirectly reachable. Although the target lies in free space, continuing along the most direct exploratory route leads into a dead-end branch. The planner therefore navigates to a different frontier to resume progress toward $\pbf_t$, ultimately enabling target reachability. Fig.~\ref{fig:aerial_robot_target_reach}c shows Mission~3, where the target is unreachable from the robot’s connected free space. In this case, the planner advances toward the closest attainable region and terminates when no further collision-free path can reduce the distance to $\pbf_t$, reporting completion once the robot reaches the closest achievable configuration within $\rho_{\mathrm{reach}}$ of the target.

        \subsubsection{\textbf{Ground Robot: Exploration Behavior}}
            \begin{figure}[t]
                \centering
                \includegraphics[clip, trim = 0cm 0cm 0cm 0cm, width=1\linewidth]{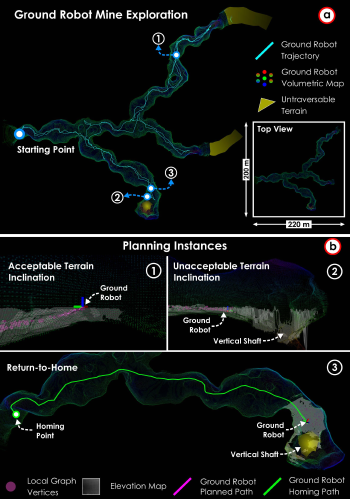}
                \caption{Simulation experiment for ground robot exploration in a large-scale mine tunnel network. (a) Full volumetric map reconstructed by the ground robot, showing the executed trajectory, untraversable terrain regions, and representative planning events during exploration. (b) Key planning instances illustrating terrain-aware sampling, where configurations on acceptable terrain inclinations are retained while samples on excessively steep or unsupported terrain are rejected based on elevation-map constraints. The figure also demonstrates return-to-home behavior, in which the planner computes a safe homing path through the explored mine structure.}
                \label{fig:ground_robot_sim}
            \end{figure}
            We demonstrate ground robot exploration using \acs{omniplanner} in a large-scale mine tunnel network environment (Fig.~\ref{fig:ground_robot_sim}). The robot operates in initially unknown terrain while incrementally constructing both a volumetric map $\mathcal{M}$ and an elevation map $\mathcal{H}$ used to enforce terrain feasibility constraints. At each planning iteration,  the Local Planning Module generates collision-free motions that satisfy terrain inclination and embodiment constraints. In this mission, the robot traverses a total path length of $1260.9$~m over $736.8$~s.

            Fig.~\ref{fig:ground_robot_sim}a shows the full exploration trajectory in a mine environment spanning approximately $220 \times 200$~m. The robot incrementally maps the tunnel network while avoiding untraversable regions identified from elevation data. Representative planning events are highlighted to illustrate how the planner selects feasible frontier directions in branching tunnel structures. The resulting trajectory demonstrates sustained exploration across multiple tunnel segments without entering terrain that violates slope or support constraints.

            Fig.~\ref{fig:ground_robot_sim}b.1-b.2 presents selected local planning instances illustrating terrain-aware sampling. Candidate configurations are projected onto the elevation map $\mathcal{H}$, and samples whose projected footprint violates the maximum allowable slope $\theta_{\max}$ or lacks terrain support are rejected. This mechanism prevents the planner from proposing motions across excessively steep or unsupported terrain while maintaining connectivity along feasible corridors. 

            Finally, Fig.~\ref{fig:ground_robot_sim}b.3 demonstrates global repositioning and return-to-home behavior. When the time budget runs out, the planner computes a safe homing path along $\Gbb_G$, allowing the robot to reliably return to the start location through previously validated terrain.

         \subsubsection{\textbf{Underwater Robot: Exploration-Inspection Behavior}}
            \begin{figure}[t]
                \centering
                \includegraphics[clip, trim = 0cm 0cm 0cm 0cm, width=1\linewidth]{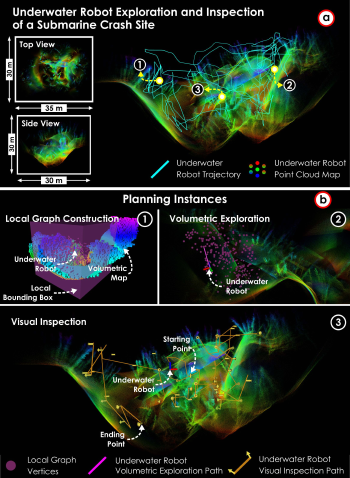}
                \caption{Simulation experiment for underwater robot exploration and inspection of a submarine crash-site. (a) Point-cloud map reconstructed during the mission, showing the executed trajectory in a $35 \times 30 \times 30$~m environment. (b) Representative planning instances illustrating local graph construction within the bounded planning volume, where sampled configurations are constrained to remain close to observed structure. The figure also shows examples of the volumetric exploration and visual inspection behaviors, in which the planner selects informative paths to expand into previously unseen regions and inspection viewpoints along the structure to maximize surface coverage.}
                \label{fig:underwater_robot_sim}
            \end{figure}
            
            We demonstrate underwater exploration and inspection using \acs{omniplanner} in a simulated submarine crash-site environment (Fig.~\ref{fig:underwater_robot_sim}). The robot operates in initially unknown space while incrementally constructing a volumetric map $\mathcal{M}$ used for collision avoidance and inspection planning. At each planning iteration, the Local Planning Module generates collision-free motions subject to embodiment constraints, while the inspection objective selects informative viewpoints to maximize surface coverage. In this mission, the robot traveled a total path length of $973.9$~m over $1401.9$~s.

            Fig.~\ref{fig:underwater_robot_sim}a shows the reconstructed point-cloud map of the crash-site from top and side views, along with the executed trajectory. The planner incrementally explores the environment while maintaining proximity to observed structure to satisfy underwater sensing constraints. The resulting trajectory demonstrates sustained exploration across complex geometry without entering large open-water regions.

            Fig.~\ref{fig:underwater_robot_sim}b presents representative planning instances. It first illustrates local graph construction within the bounded planning volume, where collision-free configurations are sampled near observed structure and connected through admissible edges to enable safe expansion in cluttered geometry. The same subfigure also shows example instances of the two behaviors considered in this scenario: (i) volumetric exploration, in which the planner selects informative paths to reduce the unseen regions, and (ii) visual inspection, in which the planner selects viewpoints along the structure to maximize surface coverage while respecting sensing range and safety constraints.
        
    \subsection{Comparison with State-of-the-Art Planning Methods}
        This subsection compares the proposed planner with state-of-the-art planning methods for aerial, ground, and underwater robots in simulation. The evaluation covers the three supported behaviors of \acs{omniplanner}, namely exploration, inspection, and target reach, using behavior-specific performance measures such as explored volume over time, inspection coverage, target-reaching performance, and aggregate efficiency metrics. These comparisons highlight the ability of \acs{omniplanner} to support multiple autonomous behaviors within a unified planning framework, while also demonstrating the distinctive capability of the inspection behavior to immediately utilize the online results of exploration.

        \subsubsection{\textbf{Simulation Setup}}
            We quantitatively compare the proposed framework against representative state-of-the-art exploration planners. The aerial and ground robot simulations are conducted in the Gazebo Classic~\cite{gazbeoclassic,2004gazebo} simulator, whereas the underwater robot simulations were done in the HoloOcean~\cite{Potokar22icraholoocean,romrell2025previewholoocean20} simulator. For each environment, all methods are initialized from the same robot configuration. A run terminates when the corresponding planner declares mission completion, which may occur due to (i) full environment exploration, (ii) planner failure or deadlock (e.g., getting stuck), or (iii) premature termination despite incomplete exploration.

            We evaluate performance across multiple simulated environments with increasing structural complexity. For aerial robots, we consider two representative settings: (i) large-scale cave environments with both multi-branch and single-branch topologies (Fig.~\ref{fig:aerial_robot_comparisons}), and (ii) a confined ballast water tank environment (Fig.~\ref{fig:aerial_comp_bwt}). For ground robots, we evaluate in a mine environment consisting of multiple traversable corridors and junctions (Fig.~\ref{fig:ground_robot_comparisons}). For the underwater robot, we conduct the simulations in a submarine crash-site. The parameters used by \acs{omniplanner} are listed in Table~\ref{tab:sim_comparative_params}.

        \subsubsection{\textbf{Baselines}}
            \begin{table*}[t!]
                \centering
                \caption{Comparison of representative planning frameworks across supported morphologies, behaviors, open-sourced code, and unified planning capability.}
                \label{tab:unification_comparison_unified}
                \renewcommand{\arraystretch}{1}
                \setlength{\tabcolsep}{14pt}
            
                \begin{threeparttable} 
                \begin{tabular}{lcccccccc}
                    \toprule
                    \multirow{2}{*}{\textbf{Method}}
                    & \multicolumn{3}{c}{\textbf{Morphology}$^1$}
                    & \multicolumn{3}{c}{\textbf{Behavior}$^2$}
                    & \multirow{2}{*}{\textbf{\shortstack{Open-sourced$^3$\\Code}}}
                    & \multirow{2}{*}{\textbf{Unified}$^4$} \\
            
                    \cmidrule(lr){2-4}
                    \cmidrule(lr){5-7}
            
                    & Aerial & Ground & Underwater
                    & VE & VI & TR
                    &
                    & \\
                    \midrule
            
                    NBVP     
                    & \CIRCLE & \LEFTcircle & \LEFTcircle
                    & \cmark & \xmark & \xmark
                    & A 
                    & \xmark\\
        
                    TARE     
                    & \CIRCLE & \CIRCLE & \Circle
                    & \cmark & \xmark & \xmark
                    & G
                    & \xmark\\
        
                    FUEL     
                    & \CIRCLE & \Circle & \Circle
                    & \cmark & \xmark & \xmark
                    & A
                    & \xmark\\
        
                    DSVP     
                    & \Circle & \CIRCLE & \Circle
                    & \cmark & \xmark & \xmark
                    & G
                    & \xmark\\
        
                    ERRT     
                    & \CIRCLE & \Circle & \Circle
                    & \cmark & \xmark & \xmark
                    & A
                    & \xmark\\
        
                    FAR Planner     
                    & \CIRCLE & \CIRCLE & \Circle
                    & \xmark & \xmark & \cmark
                    & G
                    & \xmark\\

                    FC-Planner     
                    & \CIRCLE & \Circle & \Circle
                    & \xmark & \cmark & \xmark
                    & A
                    & \xmark\\
        
                    GBPlanner 2.0     
                    & \CIRCLE & \CIRCLE & \Circle
                    & \cmark & \xmark & \xmark
                    & A$|$G
                    & \xmark\\
        
                    \midrule
            
                    \acs{omniplanner}$^\ast$     
                    & \CIRCLE & \CIRCLE & \CIRCLE
                    & \cmark & \cmark & \cmark
                    & A$|$G$|$U
                    & \cmark\\
                    
                    \bottomrule
                \end{tabular}
                
                \vspace{2pt}
                \begin{tablenotes}
                    \footnotesize{
                        \item $^{1}$ Supported robot morphologies, where $\CIRCLE$: directly supported by the original work, $\LEFTcircle$: supported only through independent extensions, adaptations, or modified variants, and $\Circle$: not demonstrated or not claimed. 
                        
                        \item $^{2}$ Planning behaviors, where \textbf{VE}: volumetric exploration, \textbf{VI}: visual inspection, and \textbf{TR}: target reach. 
                        
                        \item $^{3}$ Open-sourced code support, where \textbf{A}: aerial, \textbf{G}: ground, and \textbf{U}: underwater. Multiple letters indicate that the released implementation supports multiple morphologies. 
                        
                        \item $^{4}$ Unified planning support, where \cmark: a single planning framework supports all considered morphologies and behaviors using the same planning core, and \xmark: no such unified support is demonstrated. Independent morphology-specific extensions are not considered evidence of unified support.

                        \item $^\ast$ The \acs{omniplanner} implementation will be released upon acceptance. 
                    }
                \end{tablenotes}
                \end{threeparttable} 
            \end{table*}
            \begin{table*}[t]
                \centering
                \caption{Comparative study parameters}
                \label{tab:sim_comparative_params}
                \setlength{\tabcolsep}{10.0pt}
                \begin{tabular}{lcccc}
                    \toprule
                    \textbf{Parameter}          & \textbf{Aerial: Cave} & \textbf{Aerial: Confined} & \textbf{Ground: Mine} & \textbf{Underwater: Submarine Crash-Site} \\
                    \hline
                    $F^{\Ds}_H, F^{\Ds}_V$      & $360^\circ, 90^\circ$ & $100^\circ, 70^\circ$ & $360^\circ, 90^\circ$ & $90^\circ, 90^\circ$ \\
                    $r_V$                       & $\SI{0.2}{\meter}$    & $\SI{0.2}{\meter}$    & $\SI{0.3}{\meter}$ & $\SI{0.4}{\meter}$ \\
                    $r_H$                       & -                     & -                     & $\SI{0.2}{\meter}$ & - \\
                    $d^x_h, d^y_h$              & -                     & -                     & $\SI{40}{\meter}, \SI{40}{\meter}$ & - \\
                    $\Bbf_R$ (m)                & $[0.6, 0.6, 0.6]$              & $[0.3, 0.3, 0.2]$     & $[0.4, 0.4, 0.2]$ & $[1.0, 1.0, 1.0]$ \\
                    $h_0$                       & -                     & -                     & $\SI{0.75}{\meter}$ & - \\
                    Sampling distribution       & \texttt{Uniform}      & \texttt{Uniform}      & \texttt{Uniform} & \texttt{Uniform} \\
                    Graph construction          & \texttt{Incremental}        & \texttt{Batch}        & \texttt{Batch} & \texttt{Incremental} \\
                    $n^{\Vs}_{\max}$            & $400$                 & $500$                 & $400$ & $400$ \\
                    $n^{\Es}_{\max}$            & $7000$                & $17000$               & $14000$ & $7000$ \\
                    $\mu_l$                      & $0.01$                & $0.05$               & $0.1$ & $0.01$ \\
                    $\mu_d$                      & $0.05$                & $0.01$               & $0.15$ & $0.15$ \\
                    \bottomrule
                \end{tabular}
            \end{table*}

            We include the following baselines to cover complementary exploration paradigms and to enable comparison against established methods for each embodiment.

            \begin{itemize}
                \item \textbf{NBVP}~\cite{bircher2016receding}: A receding-horizon next-best-view exploration planner that incrementally selects informative viewpoints based on expected information gain. We include NBVP because it has been extensively used as a benchmark in aerial exploration and has also been adopted for underwater exploration tasks~\cite{2020sureshuwnbvp}, enabling comparative evaluation across both aerial and underwater robotic platforms.

                \item \textbf{TARE}~\cite{cao2021tare}: An exploration framework that integrates local planning with global reasoning and revisit mechanisms to improve long-range exploration efficiency in large environments. We include TARE as a widely adopted baseline in the ground-robot autonomous exploration benchmark and for its demonstrated performance in underground exploration scenarios.

                \item \textbf{FUEL}~\cite{zhou2021fuel}: A viewpoint-based exploration method that samples candidate viewpoints, evaluates them using a gain--cost utility (expected newly observed volume versus travel cost), and replans online as the map is updated. We include FUEL as a representative utility-driven viewpoint selection strategy and to evaluate performance under the constrained field-of-view and visibility limitations characteristic of confined-space exploration environments.

                \item \textbf{DSVP}~\cite{zhu2021dsvp}: An exploration planner based on decision-space and viewpoint reasoning that aims to improve coverage and navigation efficiency in cluttered environments. We include DSVP as a representative baseline in the ground-robot autonomous exploration benchmark, providing a complementary exploration strategy to TARE for evaluating autonomous exploration performance.
                
                \item \textbf{ERRT}~\cite{lindqvist2024tree}: A purely local exploration planner for aerial robots that expands a rapidly-exploring random tree in the robot’s reachable space and selects motions based on local information gain. We include ERRT as a representative local-only baseline to assess performance in the absence of explicit global memory or repositioning.

                \item \textbf{FAR Planner}~\cite{yang2022far}: A target reach planning framework based on a dynamically maintained visibility graph that incrementally captures the navigable structure of the environment as new observations become available. The planner performs global route optimization over the visibility graph while supporting rapid replanning and recovery from dead ends in unknown environments. We include FAR Planner as a representative state-of-the-art target reach baseline and to evaluate the performance of our approach against a specialized navigation planner designed specifically for efficient goal-directed motion in partially known environments.

                \item \textbf{FC-Planner}~\cite{feng2024fc}: An aerial coverage planner that uses skeleton-guided decomposition, informative viewpoint generation, and hierarchical path planning to inspect complex 3D structures. We select FC-Planner as the inspection baseline because it is specifically designed for UAV-based coverage and visual inspection of complex structures, making it a relevant comparison for the industrial inspection scenario.
                 
                \item \textbf{GBPlanner~2.0}~\cite{kulkarni2022autonomous}: An exploration planner based on a bifurcated local--global planning architecture that couples a local volumetric exploration layer with a global graph for frontier-based repositioning. We include GBPlanner~2.0 because it is a widely used benchmark in both aerial and ground autonomous exploration studies and represents a state-of-the-art local--global exploration strategy closely related to our graph-based approach.         
            \end{itemize}
            
            Before presenting the quantitative comparisons, we first provide a feature comparison in Table~\ref{tab:unification_comparison_unified}, which summarizes, to the best of our knowledge, the scope of representative planning frameworks considered in this work. The table distinguishes direct morphology support by the original work from support demonstrated only through independent extensions, adaptations, or modified variants. It also reports the planning behaviors addressed by each method and the robot morphologies supported by the corresponding open-sourced implementation, when available. Importantly, the unified planning column is reserved for methods that provide a single planning framework across all considered morphologies, i.e., aerial, ground, and underwater robots, using the same planning core. Therefore, independent morphology-specific extensions of an existing method are not considered evidence of unified support.

        \subsubsection{\textbf{Aerial Robot Evaluation}}
            \begin{figure*}[t!]
                \centering
                \subfloat[Multi-branch cave environment]{
                    \includegraphics[height=4.6cm]{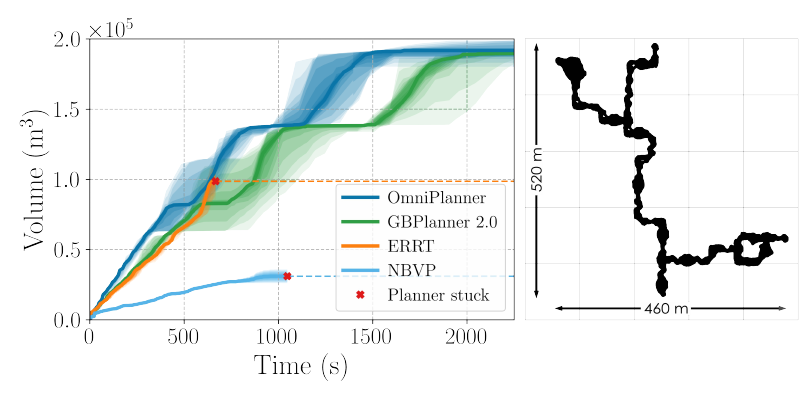}
                    \label{fig:aerial_comp_full_map_env}
                }
                \hfill
                \subfloat[Single-branch cave environment]{
                    \includegraphics[height=4.6cm]{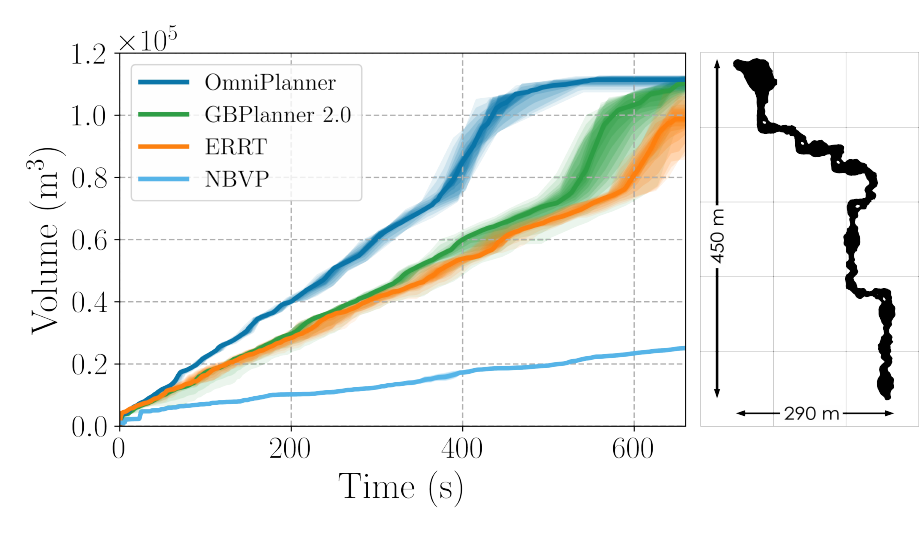}
                    \label{fig:aerial_comp_single_branch_env}
                }
                \caption{Aerial robot exploration performance of \acs{omniplanner}, GBPlanner~2.0, ERRT, and NBVP in cave environments of different structural complexity over 10 independent runs. The explored volume is reported as a function of time. Solid lines denote the median performance across runs, and shaded regions indicate the 5th–95th percentile range. (a) Multi-branch cave environment. \acs{omniplanner} and GBPlanner~2.0 successfully explore the entire environment, whereas ERRT becomes trapped in a local region and fails to continue exploration. NBVP also exhibits limited exploration progress and does not achieve full coverage. (b) Single-branch cave environment. The simplified topology reduces the need for global repositioning, enabling \acs{omniplanner}, GBPlanner~2.0, and ERRT to achieve comparable final coverage, although \acs{omniplanner} reaches high coverage more rapidly. NBVP consistently explores a substantially smaller volume than the other methods.}
                \label{fig:aerial_robot_comparisons}
            \end{figure*}

            \begin{table*}
                \caption{Aerial robot evaluation in a large-scale environment.}
                \label{tab:aerial_robot_comparison_single_branch_cave}
                \centering
                \setlength{\tabcolsep}{10.0pt}
                \renewcommand{\arraystretch}{1.2}
                \begin{tabular}{lccccc}
                    \toprule
                    \textbf{Metric} & \textbf{ERRT} & \textbf{FUEL} & \textbf{NBVP} & \textbf{GBPlanner~2.0} & \textbf{\acs{omniplanner}} \\
                    \hline
                    Explored Volume (\%) $\uparrow$ & $85.63\pm5.46$ & $\times$ & $26.98\pm6.18$ & $\underline{96.05\pm2.74}$ & $\mathbf{96.56\pm1.53}$\\
                    Time--to--$50\%$ Volumetric Coverage (s) $\downarrow$  & $439.54\pm16.83$ & $\times$ & $\times$ & $\underline{402.88\pm27.49}$ & $\mathbf{295.25\pm14.76}$\\
                    Time--to--$95\%$ Volumetric Coverage (s) $\downarrow$  & $\times$ & $\times$ & $\times$ & $\underline{665.25\pm31.71}$ & $\mathbf{519.22\pm35.94}$\\
                    Computation Time (ms) $\downarrow$   & $1362.60$ & $\times$ & $4418$ & $\underline{210.60}$ & $\mathbf{153.80}$\\
                    \bottomrule
                \end{tabular}\\[0.5ex]
                \parbox{0.95\linewidth}{\footnotesize $^*$ Best results are shown in \textbf{bold} and second-best results are \underline{underlined}.}
            \end{table*}            

            To evaluate \acs{omniplanner} for aerial exploration, we conduct simulation studies in both large-scale cave environments and confined industrial-like environments. The evaluation includes ERRT, NBVP, GBPlanner 2.0, and FUEL where applicable, covering both local receding-horizon planners and local–global exploration frameworks. All methods are evaluated using identical robot dimensions, sensing configurations, initial conditions, and computational resources. Performance is reported using final explored volume, time-to-50$\%$ volumetric coverage, time-to-95$\%$ volumetric coverage, and average computation time per planning cycle.
            
            \mypar{Large-scale Environment} We conducted simulation-based comparisons in the large-scale cave benchmark under two settings: (i) a multi-branch cave spanning approximately $520 \times 460$~m and (ii) a single-branch cave spanning approximately $450 \times 290$~m, as shown in Fig.~\ref{fig:aerial_robot_comparisons}. For each environment, all methods were evaluated over $10$ independent trials using identical initial conditions and planning budgets. In this study, a $3$D LiDAR sensor model with $[F^{\Ds}_H, F^{\Ds}_V] = [360^{\circ}, 90^{\circ}]$ and $d^{\Ds}_{\max} = 20$ m was used as the depth sensor.

            Fig.~\ref{fig:aerial_robot_comparisons} reports the explored volume as a function of time, where solid curves indicate the median performance across runs and shaded regions denote the $5^{\mathrm{th}}$--$95^{\mathrm{th}}$ percentile range. In the large-scale multi-branch cave environment (Fig.~\ref{fig:aerial_robot_comparisons}a), \acs{omniplanner} achieves the strongest overall performance, demonstrating the benefits of combining persistent global connectivity with continuous local replanning. Similar to \acs{omniplanner}, GBPlanner~2.0 maintains exploration progress through global repositioning, highlighting the effectiveness of local--global exploration architectures. However, its more conservative planning--execution cycle results in slower exploration progress. In contrast, ERRT initially explores effectively through local information-gain optimization but eventually stagnates in locally informative regions, while NBVP exhibits a more pronounced degradation due to the scalability limitations of its single-tree receding-horizon formulation in large, branching environments where long-range repositioning becomes necessary.
            
            To isolate the effect of the local exploration strategy from the benefits of global repositioning, we additionally evaluate a single-branch cave environment (Fig.~\ref{fig:aerial_robot_comparisons}b), where exploration naturally progresses along a single cave branch. This setting provides a more equitable comparison with ERRT and NBVP, allowing the observed performance differences to be attributed primarily to the effectiveness of the local planning and viewpoint-selection strategies rather than to the availability of global memory. FUEL was also considered for this benchmark, however, its point-cloud-based representation could not be executed reliably at the scale and map resolution required by the cave environment due to memory limitations. Reducing the map density sufficiently to fit within the available resources resulted in non-representative exploration behavior.

            \begin{table*}[t]
                \caption{Aerial robot evaluation in a confined environment.}
                \label{tab:aerial_robot_comparison_bwt}
                \centering
                \setlength{\tabcolsep}{8.0pt}
                \renewcommand{\arraystretch}{1.2}
                \begin{tabular}{lccccc}
                    \toprule
                    \textbf{Metric} & \textbf{ERRT} & \textbf{FUEL} & \textbf{NBVP} & \textbf{GBPlanner~2.0} & \textbf{\acs{omniplanner}} \\
                    \hline
                    Explored Volume (\%) $\uparrow$ & $63.23\pm12.56$ & $95.18\pm5.03$ & $77.4\pm13.3$ & $\underline{95.34\pm2.47}$ & $\mathbf{98.7\pm0.22}$\\
                    Time--to--$50\%$ Volumetric Coverage (s) $\downarrow$  & $\underline{104.53\pm23.27}$ & $241.81\pm33.54$ & $247.95\pm5.08$ & $168.6\pm66.05$ & $\mathbf{72.62\pm9.07}$\\
                    Time--to--$95\%$ Volumetric Coverage (s) $\downarrow$ & $\times$ & $\underline{461.03\pm99.45}$ & $\times$ & $560\pm42.46$ & $\mathbf{361.2\pm56.41}$\\
                    Computation Time (ms) $\downarrow$ & $3121.90$ & $\mathbf{54.39}$ & $5236$ & $312.23$ & $\underline{278.15}$\\
                    \bottomrule
                \end{tabular}\\[0.5ex]
                \parbox{0.95\linewidth}{\footnotesize $^*$ Best results are shown in \textbf{bold} and second-best results are \underline{underlined}.}
            \end{table*}
            
            The quantitative comparison of exploration efficiency and completion time is reported for the single-branch cave environment in Table~\ref{tab:aerial_robot_comparison_single_branch_cave}. Even in the absence of a significant global repositioning requirement, \acs{omniplanner} achieves the best overall performance. This indicates that its advantage is not solely derived from the global planning architecture, but also from the efficiency of its local planning and execution framework. In particular, the Local Planning Module is triggered proactively while the robot is still traversing the current path segment, reducing idle time and avoiding the stop--plan--go behavior adopted by competing methods. As a result, the robot spends a larger fraction of the mission actively exploring rather than waiting for planning updates. Furthermore, the lower planning overhead of \acs{omniplanner} enables more frequent replanning and faster reaction to newly observed information, leading to more efficient viewpoint selection and exploration progress. These results suggest that, even in environments where global reasoning provides limited benefit, the tighter integration between planning and execution remains a key contributor to the superior performance of \acs{omniplanner}.

            \begin{figure}[t]
                \centering
                \includegraphics[clip, trim = 0.5cm 0cm 0cm 0cm, width=1\linewidth]{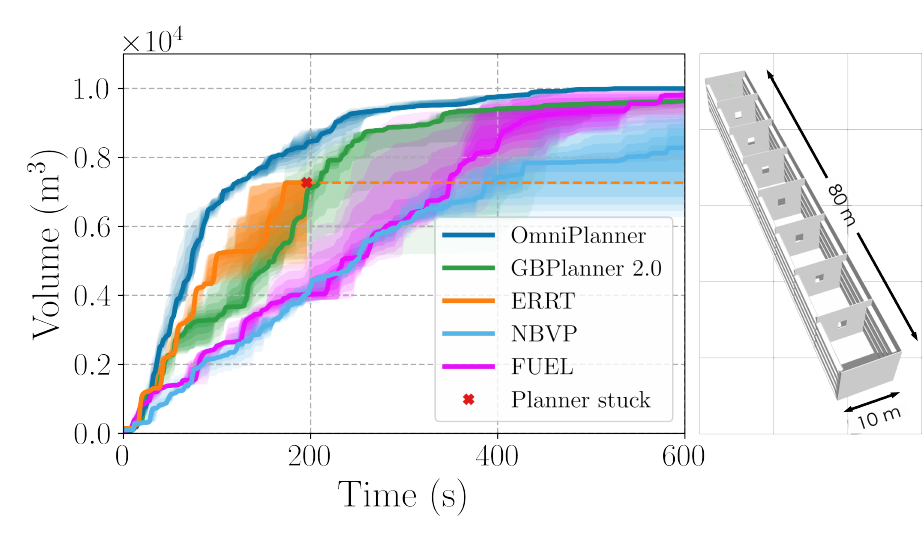}
                \caption{Aerial robot exploration performance of \acs{omniplanner}, GBPlanner~2.0, ERRT, NBVP, and FUEL in a custom-designed ballast water tank comprising 8 compartments connected by manholes (openings). Results are aggregated over 5 independent runs. The explored volume is reported as a function of time. Solid curves represent median performance, and shaded regions indicate the 5th–95th percentile range. The red marker denotes the point at which ERRT becomes trapped and can no longer increase explored volume.}
                \label{fig:aerial_comp_bwt}
            \end{figure}
            
            \mypar{Confined Environment} In contrast with the large-scale environment, a custom-designed confined ballast water tank comprising eight compartments connected through narrow manholes (openings), as shown in Fig.~\ref{fig:aerial_comp_bwt}, is used to evaluate the aerial exploration performance. The environment spans approximately $80 \times 10$~m and represents a highly constrained setting where successful exploration requires both efficient viewpoint selection and reliable traversal through narrow inter-compartment passages. All planners were initialized from the same starting location and evaluated over five independent trials using identical planning budgets. In this study, a depth camera sensor model with $[F^{\Ds}_H, F^{\Ds}_V] = [100^{\circ}, 70^{\circ}]$ and $d^{\Ds}_{\max} = 7.5$ m was used as the depth sensor.

            Fig.~\ref{fig:aerial_comp_bwt} reports the explored volume as a function of time, where solid curves indicate the median performance across runs and shaded regions denote the $5^{\mathrm{th}}$--$95^{\mathrm{th}}$ percentile range. Table~\ref{tab:aerial_robot_comparison_bwt} summarizes the exploration performance in the ballast water tank environment. \acs{omniplanner} achieves the highest final coverage ($98.7\pm0.22\%$), demonstrating both efficient and reliable exploration. While GBPlanner~2.0 and FUEL achieve comparable final coverage, they require substantially longer exploration times, whereas NBVP and ERRT fail to sufficiently explore the entire environment.

            The superior performance of \acs{omniplanner} can be attributed to its batch graph construction strategy, introduced in Section~\ref{sec:planning_kernel}.1, which enables more efficient graph expansion and connectivity establishment in confined environments while maintaining lower computational cost. By rapidly discovering and connecting transitions between compartments, \acs{omniplanner} achieves faster coverage growth and reaches unexplored regions more effectively than competing graph-based approaches. In contrast, GBPlanner~2.0 relies on a less efficient graph construction process, resulting in increased planning overhead and slower long-range exploration. Although FUEL eventually attains comparable final coverage, its frontier-based exploration strategy tends to exhaustively cover individual compartments before transitioning to neighboring regions, leading to slower exploration progress in confined environments.

            ERRT exhibits rapid initial exploration but frequently becomes trapped after entering a compartment. This behavior is largely due to the planner's original design for omnidirectional sensing systems, as it does not explicitly reason about viewpoint yaw. Under the constrained FoV sensing configuration employed in this benchmark, successful traversal of openings requires deliberate viewpoint orientation, limiting ERRT's ability to consistently progress between compartments. NBVP similarly struggles to efficiently explore the environment due to its receding-horizon tree exploration strategy, which tends to focus on locally informative regions and lacks an effective mechanism for long-range repositioning.

            \begin{figure}[t!]
                \centering
                \includegraphics[clip, trim = 0.0cm 0cm 0cm 0cm, width=1\linewidth]{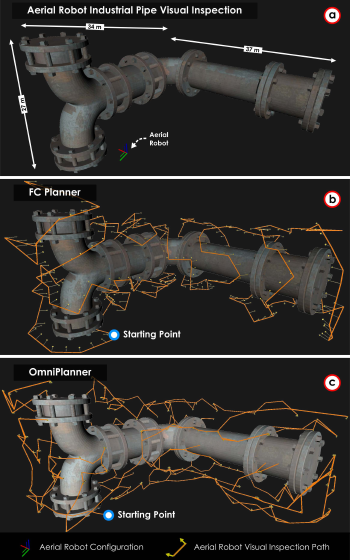}
                \caption{\textbf{Aerial robot visual inspection behavior comparison in an industrial pipe environment.} Inspection paths generated by (b) FC-Planner and (c) \acs{omniplanner} for sequential visual inspection of an industrial pipe. Starting from the marked initial position, the robot navigates around the pipe structure while adjusting its actuated camera to inspect the exterior surface. Both planners successfully complete the inspection mission.}
                \label{fig:inspection_aerial_robot_comparion_path}
            \end{figure}
            
            \textbf{Visual Inspection Behavior Comparison.} To evaluate the visual inspection behavior of \acs{omniplanner}, we compare it against FC-Planner in an industrial pipe inspection scenario representative of structured infrastructure monitoring. FC-Planner is selected as the baseline because it is a recent specialized aerial coverage/inspection planner designed for efficient surface coverage in complex 3D environments, making it a relevant task-specific reference for evaluating the inspection behavior of \acs{omniplanner}. The environment consists of a large pipe assembly with long cylindrical segments, flanges, bolts, and curved junctions, which creates a challenging inspection setting with viewpoint-dependent visibility constraints. In this task, the robot must not only reach informative viewpoints around the structure, but also maintain collision-free motion while selecting camera poses that provide sufficient surface coverage. This scenario is therefore well suited for assessing whether a planner can produce efficient inspection paths around complex three-dimensional industrial assets. For both methods, the aerial robot is initialized from the same starting location and is tasked with inspecting the exterior surface of the pipe using an actuated camera. The comparison focuses on the resulting inspection path, achieved surface coverage, number of selected viewpoints, path length, and computation time.
            
            Fig.~\ref{fig:inspection_aerial_robot_comparion_path} and Table~\ref{tab:aerial_robot_comparison_pipe} compare the inspection behavior of FC-Planner and \acs{omniplanner} in the industrial pipe scenario, with the quantitative results reported over five trials. Both planners complete the mission and achieve near-complete surface coverage, but their paths exhibit different characteristics. FC-Planner follows the pipe geometry through a more distributed set of viewpoints, while \acs{omniplanner} generates a more compact path with smoother transitions between inspection viewpoints. This trend is consistent with the design of the two methods: FC-Planner relies on skeleton-guided decomposition and specialized viewpoint generation for aerial coverage planning, whereas \acs{omniplanner} selects coverage-informative viewpoints directly within the same local graph used for motion feasibility. This coupling helps avoid spatially inefficient viewpoints, while the actuated camera increases the feasible viewing directions from each sampled position and reduces redundant robot motion. Consequently, \acs{omniplanner} achieves comparable inspection completeness to a specialized coverage-planning baseline, with slightly better trends in path compactness, viewpoint efficiency, and computation time, while using the same unified framework employed for exploration and target reach behaviors.

            \begin{table}[t!]
                \caption{Aerial robot evaluation performing visual inspection of an industrial pipe.}
                \label{tab:aerial_robot_comparison_pipe}
                \centering
                \setlength{\tabcolsep}{7.0pt}
                \renewcommand{\arraystretch}{1.2}
                \begin{tabular}{lcc}
                    \toprule
                    \textbf{Metric} & \textbf{FC-Planner} & \textbf{\acs{omniplanner}} \\
                    \hline
                    Covered Surface (\%) $\uparrow$ & $97.36\pm0.49$ & $\mathbf{97.82\pm0.93}$\\
                    Viewpoints (\#) $\downarrow$ & $188.67\pm7.23$ & $\mathbf{186.94\pm5.15}$\\
                    Path Length (m) $\downarrow$ & $1343.99\pm72.23$ & $\mathbf{1297.41\pm34.62}$\\
                    Computation Time (ms)   & $1804.18$ & $\mathbf{1346.01}$\\
                    \bottomrule
                \end{tabular}\\[0.5ex]
                \parbox{0.98\linewidth}{\footnotesize $^*$ Best results are shown in \textbf{bold}.}
            \end{table}

            \begin{table*}[t!]
                \caption{Ground robot evaluation in a mine environment.}
                \label{tab:ground_robot_comparison_mine}
                \centering
                \setlength{\tabcolsep}{12.0pt}
                \renewcommand{\arraystretch}{1.2}
                \begin{tabular}{lcccc}
                    \toprule
                    \textbf{Metric} & \textbf{DSVP} & \textbf{TARE} & \textbf{GBPlanner~2.0} & \textbf{\acs{omniplanner}} \\
                    \hline
                    Explored Volume (\%) $\uparrow$ & $96.01\pm3.69$ & $96.13\pm1.87$ & $\underline{96.64\pm2.93}$ & $\mathbf{97.01\pm0.24}$\\
                    Time--to--$50\%$ Volumetric Coverage (s) $\downarrow$  & $915.19\pm98.14$ & $842.01\pm111.48$ & $\underline{762.69\pm59.52}$ & $\mathbf{734.9\pm34.56}$ \\
                    Time--to--$95\%$ Volumetric Coverage (s) $\downarrow$  & $2678.61\pm108.57$ & $\underline{2259.85\pm124.19}$ & $2421.9\pm450.07$ & $\mathbf{1853.26\pm320.85}$\\
                    Computation Time (ms) $\downarrow$   & $410.80$ & $364.24$ & $\underline{111.81}$ & $\mathbf{74.57}$\\
                    \bottomrule
                \end{tabular}\\[0.5ex]
                \parbox{0.98\linewidth}{\footnotesize $^*$ Best results are shown in \textbf{bold} and second-best results are \underline{underlined}.}
            \end{table*}

        \subsubsection{\textbf{Ground Robot Evaluation}}
\begin{figure}[t]
                \centering
                \includegraphics[clip, trim = 0.5cm 0cm 0cm 0cm, width=1\linewidth]{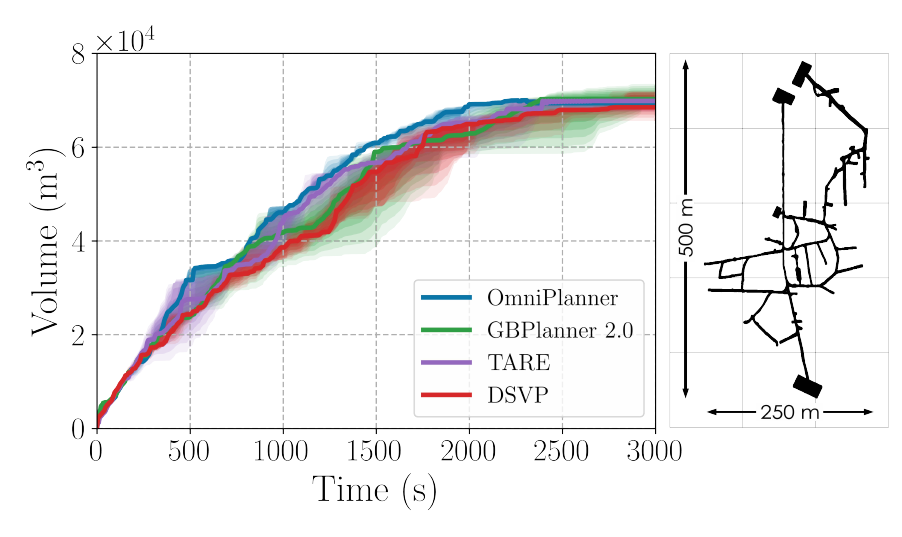}
                \caption{Ground robot exploration performance of \acs{omniplanner}, GBPlanner~2.0, TARE, and DSVP in a mine environment over 5 independent runs. The explored volume is reported as a function of time. Solid curves denote the median performance across runs, and shaded regions indicate the 5th–95th percentile range. All methods achieve similar final exploration coverage, with differences primarily in the rate at which coverage is accumulated.}
                \label{fig:ground_robot_comparisons}
            \end{figure}

            \textbf{Exploration Behavior Comparison.} We compare \acs{omniplanner} against DSVP, TARE and GBPlanner~2.0 to evaluate ground robot exploration performance in a large-scale mine environment. The environment spans approximately $500 \times 250$~m, as shown in Fig.~\ref{fig:ground_robot_comparisons}. All methods are evaluated over $5$ independent trials using identical initial conditions and planning budgets. In this study, a $3$D LiDAR sensor model with $[F^{\Ds}_H, F^{\Ds}_V] = [360^{\circ}, 90^{\circ}]$ and $d^{\Ds}_{\max} = 20$ m was used as the depth sensor. To enable a fair comparison, the Gazebo Classic-based Autonomous Exploration Development Environment~\cite{cao2022cmuenv} was used for this evaluation.

            Fig.~\ref{fig:ground_robot_comparisons} reports the explored volume as a function of time, where solid curves indicate the median performance across runs and shaded regions denote the $5^{\mathrm{th}}$--$95^{\mathrm{th}}$ percentile range. Table~\ref{tab:ground_robot_comparison_mine} summarizes the quantitative results in the mine exploration environment. All methods achieve similar final exploration coverage, with \acs{omniplanner} attaining the highest value ($97.01 \pm 0.24\%$). However, the primary distinction between the methods lies in the rate at which coverage is accumulated. \acs{omniplanner} reaches $50\%$ coverage in $734.9 \pm 34.56$~s and $95\%$ coverage in $1853.26 \pm 320.85$~s, outperforming DSVP, TARE, and GBPlanner~2.0 in both early-stage and long-range exploration efficiency.

            The superior performance of \acs{omniplanner} is primarily attributed to its unified graph representation and batch graph construction strategy, which enable efficient graph expansion and connectivity establishment across the large-scale branching mine network. Compared to GBPlanner~2.0, the proposed graph construction approach reduces planning overhead while allowing frontier information and long-range connectivity to be propagated more efficiently throughout the graph. As a result, \acs{omniplanner} is able to identify and reach unexplored regions more rapidly, leading to faster convergence despite both methods employing graph-based exploration strategies.

            TARE achieves comparable exploration progress during portions of the mission due to its hierarchical local-global exploration framework. However, its separation between local exploration and global route planning results in less efficient long-range decision making as the explored area grows, leading to slower convergence to near-complete coverage. Similarly, DSVP successfully explores the environment but requires longer exploration times because its viewpoint selection strategy is primarily driven by local exploration utility and does not explicitly leverage a persistent graph structure for efficient global repositioning.

            The computational results further highlight the scalability of the proposed approach. \acs{omniplanner} achieves the lowest average planning time ($74.57$~ms), outperforming GBPlanner~2.0 ($111.81$~ms), TARE ($364.24$~ms), and DSVP ($410.80$~ms). These results demonstrate that \acs{omniplanner} not only achieves the fastest exploration rates but also maintains the lowest computational burden.

            \textbf{Target Reach Behavior Comparison.} To evaluate the target reach capability of \acs{omniplanner}, we compare it against FAR Planner, a state-of-the-art visibility-graph-based navigation framework specifically designed for efficient goal-directed motion in partially known and unknown environments. The evaluation is conducted in the large-scale underground tunnel environment, where a ground robot is tasked with sequentially navigating from a common start location to four user-defined target positions ($\mathbf{p}_{t,1}$--$\mathbf{p}_{t,4}$). For both planners, the robot operates without prior knowledge of the environment and incrementally constructs a representation of the traversable space during execution.

            \begin{figure}[t!]
                \centering
                \includegraphics[clip, trim = 0.0cm 0cm 0cm 0cm, width=1\linewidth]{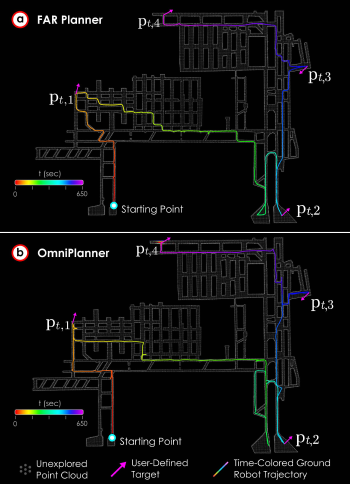}
                \caption{\textbf{Ground robot target reach behavior comparison in a large-scale underground tunnel environment.} Time-colorized ground-robot trajectories generated by (a) FAR Planner and (b) \acs{omniplanner} for sequential navigation to four user-defined target positions ($\mathbf{p}_{t,1}$--$\mathbf{p}_{t,4}$). Trajectory colors denote temporal progression from mission start to mission completion. Both planners successfully complete the target reach mission, while exhibiting different route selection strategies and traversal characteristics, especially during navigation between consecutive target positions.}
                \label{fig:target_reach_ground_robot_comparison_trajectory}
            \end{figure}
            
             Fig.~\ref{fig:target_reach_ground_robot_comparison_trajectory} illustrates representative trajectories generated by FAR Planner and \acs{omniplanner}, respectively. Both methods successfully reach all target locations. FAR Planner generally produces shorter routes owing to its visibility-graph representation and direct optimization of navigational connectivity, whereas \acs{omniplanner} employs its bifurcated local--global planning architecture to incrementally advance toward each target while simultaneously maintaining frontier information and global connectivity. Despite not being specifically designed for target reaching, \acs{omniplanner} exhibits navigation behavior qualitatively similar to FAR Planner, successfully guiding the robot through the complex tunnel network and avoiding dead ends while progressing toward the requested targets.

\begin{figure}[t!]
                \centering
                \includegraphics[clip, trim = 0.0cm 0cm 0cm 0cm, width=1\linewidth]{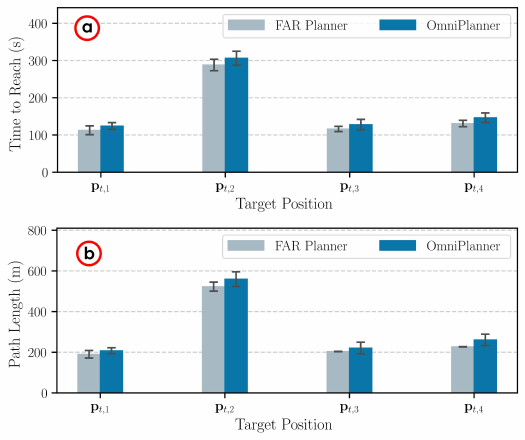}
                \caption{\textbf{Per-target navigation performance comparison.} Mean performance over five trials for FAR Planner and \acs{omniplanner} when navigating to the target positions $\mathbf{p}_{t,1}$--$\mathbf{p}_{t,4}$. (a) Average time required to reach each target position. (b) Average path length traveled to reach each target position. Error bars indicate one standard deviation across trials. The results demonstrate that \acs{omniplanner} achieves target reaching performance comparable to the specialized FAR Planner baseline while operating within a unified planning framework supporting multiple behaviors and robotic embodiments.}
                \label{fig:target_reach_ground_robot_comparison_bars}
            \end{figure}

             The quantitative results are presented in Fig.~\ref{fig:target_reach_ground_robot_comparison_bars}. Averaged over five independent trials, \acs{omniplanner} remains within $8.6\%$ of FAR Planner in mission completion time while producing trajectories that are only $7.7\%$ longer on average. The per-target results further show that the performance gap remains relatively small across all target positions, indicating that the proposed target reach behavior achieves navigation efficiency comparable to the specialized FAR Planner baseline. Considering that FAR Planner is exclusively designed for goal-directed navigation, whereas \acs{omniplanner} employs the same planning kernel for exploration, inspection, and target reach tasks across aerial, ground, and underwater robots, these results demonstrate that the proposed framework can maintain competitive target reaching performance while providing substantially broader functionality and cross-domain applicability.

        \subsubsection{\textbf{Underwater Robot Evaluation}}
            
\begin{figure}[t]
                \centering
                \includegraphics[clip, trim = 0cm 0cm 0cm 0cm, width=1\linewidth]{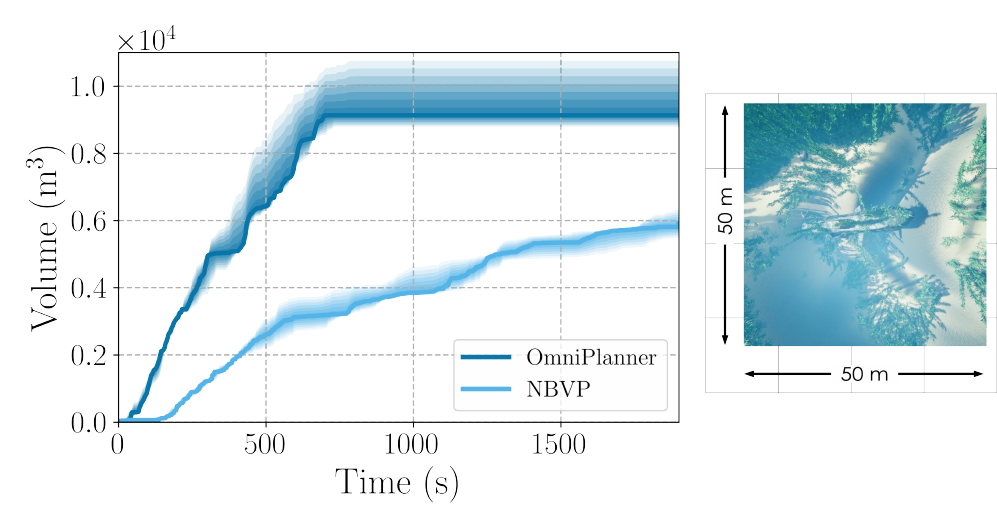}
                \caption{Underwater robot exploration performance of \acs{omniplanner} and NBVP in an underwater submarine crash-site environment over 3 independent runs. The explored volume is reported as a function of time. Solid curves represent the median performance across runs, and shaded regions indicate the 5th–95th percentile range. \acs{omniplanner} achieves higher exploration efficiency and final coverage throughout the experiment, while NBVP exhibits slower exploration progress and lower overall coverage.}
                \label{fig:underwater_robot_comparisons}
            \end{figure}

            We compare \acs{omniplanner} against NBVP, evaluating the underwater exploration performance in a simulation model of an underwater submarine crash-site. The environment contains vegetation and a crashed submarine providing additional structures. The environment spans approximately $50 \times 50 \times 25 $~m, as shown in Fig.~\ref{fig:underwater_robot_comparisons}. All methods are evaluated over 3 runs, starting at the same location. In this study, a depth camera with $[F^{\Ds}_H, F^{\Ds}_V] = [90^{\circ}, 90^{\circ}]$ and $d^{\Ds}_{\max} = 10$ m was used as the depth sensor.

            Fig.~\ref{fig:underwater_robot_comparisons} reports the explored volume as a function of time, where solid curves indicate the median performance across runs and shaded regions denote the $5^{\mathrm{th}}$--$95^{\mathrm{th}}$ percentile range. Table~\ref{tab:underwater_robot_comparison_submarine} summarizes the quantitative results in the underwater submarine crash-site environment. \acs{omniplanner} substantially outperforms NBVP across all evaluation metrics, achieving a final exploration coverage of $96.26 \pm 3.03\%$ compared to $54.4 \pm 3.82\%$ for NBVP. Furthermore, \acs{omniplanner} reaches $50\%$ coverage in $389.36 \pm 65.22$~s, approximately four times faster than NBVP ($1571.32 \pm 255.93$~s), and is the only method capable of reaching $95\%$ coverage, requiring $631.82 \pm 56.94$~s.

            The large performance gap is primarily attributable to the scalability limitations of NBVP's receding-horizon exploration strategy. As discussed previously, NBVP repeatedly constructs and evaluates a local exploration tree without maintaining a persistent global graph structure. While effective in smaller environments, this approach becomes increasingly inefficient as the explored area grows, requiring repeated expansion of large trees and providing no efficient mechanism for long-range repositioning toward distant unexplored regions. Consequently, NBVP tends to remain focused on locally informative areas and exhibits significantly slower exploration progress once nearby information gain is exhausted. These limitations are also reflected in the computational results, where NBVP requires $4436.96$~ms per planning cycle compared to only $67.59$~ms for \acs{omniplanner}.

    \section{FIELD EXPERIMENTS}
    \label{sec:field_experiments}
    
    To validate \acs{omniplanner}’s ability to autonomously execute the proposed behaviors across heterogeneous robotic morphologies and operating domains, extensive field deployments were conducted on three types of robotic platforms: two aerial robots as described below, one ground robot, and one underwater \acs{rov}. Each platform performed fully autonomous missions in its respective environment using the same planning pipeline described in Section~\ref{sec:proposed_approach}.

    \subsection{Robotic Platforms}
        \subsubsection{\textbf{Aerial Robots}}
            The aerial experiments were conducted using two flying robots, called i) \ac{ar1} and ii) \ac{ar2}. 
            
            \acs{ar1} is a resilient aerial robot designed for autonomous missions in confined and structurally complex environments \cite{petris2022rmf}. The platform is equipped with a collision-tolerant frame measuring $0.38 \times 0.38 \times 0.24$ m (L $\times$ W $\times$ H)  and has a total mass of $1.45$ kg, providing an average flight endurance of approximately $10$ minutes. The onboard sensing and compute payload --- hereafter referred to as \ac{am1} --- consists of a Khadas VIM4 single-board computer (SBC) featuring $4\times2.2$ GHz Cortex-A73 cores and $4\times2.0$ GHz Cortex-A53 cores as the compute unit. The sensing suite of \ac{am1} includes a VectorNav VN-100 IMU, an Ouster OS0-64 LiDAR (FoV: $[360^\circ \times 90^\circ]$, maximum range: $\SI{100}{\meter}$), and a Blackfly S RGB camera (FoV: $[85^\circ \times 64^\circ]$, resolution: $720 \times 540$ px). The robot runs CompSLAM \cite{shehryar2020complementary}, a multi-modal simultaneous localization and mapping (SLAM) framework that provides accurate odometry and dense mapping in real time. In addition, the proposed unified path planning pipeline is executed onboard to generate collision-free trajectories, which are subsequently tracked using the model predictive controller (MPC) proposed in \cite{kamel2017model}.

            \acs{ar2} is a collision-tolerant aerial robot designed for autonomous operation in GPS-denied and confined environments \cite{kulkarni2025unipilot}. The platform features a lightweight protective frame measuring approximately $0.52 \times 0.52 \times 0.24$ m (L $\times$ W $\times$ H) and a total mass of $1.47$ kg excluding payload. \acs{ar2}'s autonomy payload \ac{am2} is equipped with an NVIDIA Jetson Orin NX as the compute module and a multi-modal sensing suite including a RoboSense Airy dome LiDAR (FoV: $[360^\circ \times 90^\circ]$, max range: $\SI{60}{\meter}$), multiple MIPI cameras, a pmd flexx2 time-of-flight (ToF) camera, a D3 Embedded FMCW radar, and a VectorNav VN-100 IMU. The robot interfaces with a Pixracer flight controller running PX4 firmware, which tracks position and velocity setpoints generated by the onboard autonomy stack. All state estimation, mapping, planning, and control processes are executed fully onboard, enabling robust autonomous flight in perceptually degraded environments. 

            \begin{table}[t]
                \caption{Underwater robot evaluation in a submarine crash-site environment.}
                \label{tab:underwater_robot_comparison_submarine}
                \centering
                \setlength{\tabcolsep}{8.0pt}
                \renewcommand{\arraystretch}{1.3}
                \begin{tabular}{lcc}
                    \toprule
                    \textbf{Metric} & \textbf{NBVP} & \textbf{\acs{omniplanner}} \\
                    \hline
                    Explored Volume (\%) $\uparrow$ & $54.4\pm3.82$ & $\mathbf{96.26\pm3.03}$\\
                    \shortstack[l]{Time--to--$50\%$ \\ Volumetric Coverage (s)}  & $1571.32\pm255.93$ & $\mathbf{389.36\pm65.22}$\\
                    \shortstack[l]{Time--to--$95\%$ \\ Volumetric Coverage (s)} & $\times$ & $\mathbf{631.82\pm56.94}$\\
                    Computation Time (ms)   & $4436.96$ & $\mathbf{67.59}$\\
                    \bottomrule
                \end{tabular}\\[0.5ex]
                \parbox{0.98\linewidth}{\footnotesize $^*$ Best results are shown in \textbf{bold}.}
            \end{table}
        \subsubsection{\textbf{Ground Robot}}
            The ground experiments were conducted using ANYmal~\cite{anybotics_website}, hereby called \ac{gr1}, a quadruped mobile robot with dimensions of $0.93 \times 0.53 \times 0.80$ m (L $\times$ W $\times$ H), a mass of $50$ kg, and a payload capacity of up to $10$ kg. The platform provides a continuous operational endurance of approximately $1$ hour. 
            
            To demonstrate the generality of the proposed approach across heterogeneous autonomy payloads, three different sensing and computation configurations were evaluated. The first configuration employs the \acs{am2} as in \acs{ar2}, running a LiDAR–radar–inertial odometry pipeline based on \cite{nissov2024degradation,nissov2024robust}, which provides accurate and robust state estimation throughout the experiment. The second configuration uses the \ac{am3} payload, equipped with an NVIDIA Jetson AGX Orin. Its sensing suite includes an Intel RealSense D455 RGB-D camera, an Ouster OS0-64 LiDAR, and a VectorNav VN-100 IMU. For this configuration, a LiDAR–inertial odometry pipeline based on \cite{khedekar2025pg} was employed, similarly providing reliable odometry during deployment. Finally, the third configuration, \ac{am4}, consists of the base sensing and compute suite of the ANYmal robot. In \ac{am4}, the onboard computation is handled by two 8th-generation Intel Core i7 CPUs (6 cores each). The sensing suite of \ac{am4} includes a Velodyne VLP-16 LiDAR (FoV: $[360^\circ \times 30^\circ]$, maximum range: $\SI{100}{\meter}$) and six Intel RealSense depth cameras distributed around the body and primarily used for perception.
            
            In all configurations, the proposed planning method was executed entirely onboard the payload computer. The generated paths were transmitted to the robot and tracked using \acs{gr1}’s internal path-tracking controller.

        \begin{table*}[ht!]
            \centering
            \caption{Overview of field experiments, including deployed robot platforms, environmental settings, and mission behaviors and statistics.}
            \setlength{\tabcolsep}{5.0pt}
            \begin{threeparttable}
            \begin{tabular}{lcccccccccccc}
                \toprule
                \multirow{2}{*}{\textbf{Field Experiment}} 
                & \multicolumn{3}{c}{\textbf{Robot}} 
                & \multicolumn{4}{c}{\textbf{Environment}} 
                & \multicolumn{3}{c}{\textbf{Behavior$^1$}}
                & \multicolumn{2}{c}{\textbf{Mission Statistics}} \\
                
                \cmidrule(lr){2-4} 
                \cmidrule(lr){5-8} 
                \cmidrule(lr){9-11} 
                \cmidrule(lr){12-13}
                & Aerial & Ground & Underwater 
                & Indoor & Outdoor & Narrow & Wide 
                & VE & VI & TR
                & Path Length (m) & Duration (min)\\
                
                \midrule
                \multirow{2}{*}{Underground Mine}
                & \cmark & \xmark & \xmark 
                & \cmark & \xmark & \cmark & \xmark  
                & \cmark & \xmark & \xmark
                & $$168.5$$ & $4.3$ \\
        
                & \xmark & \cmark & \xmark 
                & \cmark & \xmark & \cmark & \xmark
                & \cmark & \xmark & \xmark
                & $357.7$ & $19.5$ \\
        
                \midrule
                University Campus 
                & \xmark & \cmark & \xmark  
                & \xmark & \cmark & \cmark & \cmark 
                & \cmark & \xmark & \xmark
                & $1228.6$ & $49.5$ \\
        
                \midrule
                \multirow{2}{*}{Forest}
                & \cmark & \cmark & \xmark 
                & \xmark & \cmark & \xmark & \cmark 
                & \cmark & \xmark & \xmark
                & $91.3~|~310.8$ \tnote{*} & $2.5~|~13.9$ \tnote{*} \\
        
                & \cmark & \xmark & \xmark
                & \xmark & \cmark & \xmark & \cmark
                & \xmark & \xmark & \cmark
                & $129.8$ & $5.1$ \\
        
                \midrule
                Ballast Water Tank 
                & \cmark  & \xmark & \xmark
                & \cmark & \xmark & \cmark & \xmark
                & \cmark & \cmark & \xmark
                & $45.1$ & $4.2$ \\         
        
                \midrule

                \multirow{2}{*}{Submarine Bunker }
                & \xmark & \xmark & \cmark
                & \xmark & \cmark & \xmark & \cmark 
                & \cmark & \xmark & \xmark
                & $238.9$ & $13.9$ \\

                & \xmark & \xmark & \cmark
                & \xmark & \cmark & \xmark & \cmark 
                & \cmark & \cmark & \xmark
                & $156.3$ & $8.2$ \\
                
                \bottomrule
            \end{tabular}
            \begin{tablenotes}
                \footnotesize{
                    \item $^{1}$ Planning behaviors, where \textbf{VE}: volumetric exploration, \textbf{VI}: visual inspection, and \textbf{TR}: target reach.
    
                    \item[*] Values are reported as $\mathfrak{a}~|~\mathfrak{b}$, where $\mathfrak{a}$ corresponds to the aerial robot and $\mathfrak{b}$ corresponds to the ground robot.
                   
                }
            \end{tablenotes}
            \end{threeparttable}
            \label{tab:field_experiments}
        \end{table*}
        
        \subsubsection{\textbf{Marsupial Ground-Aerial Robot Team}}\label{susubsec:marsupial_robots}
             The marsupial system comprises a heterogeneous ground-aerial robot team consisting of the \acs{gr1} quadruped ground robot carrying the \acs{am4} payload and the \acs{ar1} aerial robot carrying the \acs{am1} payload operating in a marsupial configuration inspired by our previous work \cite{zacharia2025collaborative}. \acs{ar1} is mounted on a rigid, custom-designed mechanism consisting of a flat platform with motorized brackets, powered by the ground robot, that securely holds the aerial robot during ground operation and enables controlled detachment for autonomous aerial deployment. Both robots maintain independent onboard computation, sensing, and state estimation pipelines, and communicate over a wireless network to enable bidirectional data exchange when connectivity is available. This design supports coordinated operation while allowing each robot to function autonomously after deployment, with all sensing, computation, and control processes executed fully onboard their respective autonomy modules.

        \subsubsection{\textbf{Underwater Robot}}
            The underwater experiments were conducted using \ac{ur1}, a custom modification of the BlueROV2 Heavy Configuration platform \cite{BlueROV2Heavy}. \acs{ur1} integrates the autonomy payload \ac{am5} consisting of an Alphasense Core Research Development Kit comprising five monochrome Sony IMX-287 global-shutter cameras ($0.4$ MP) rigidly mounted on a common frame (FoV: $[126^\circ \times 92.4^\circ]$). The cameras are tightly synchronized with a Bosch BMI085 IMU using a mid-frame, exposure-compensated scheme, achieving sub-100 $\mu$s synchronization accuracy. An NVIDIA Jetson AGX Orin compute board is utilized to perform all the computations onboard the robot as part of \ac{am5}, while the high-level commands and telemetry are supported by a tether cable. The robot state estimation is based on ReAqROVIO \cite{mohit2024refractive}, a refraction-aware multi-camera visual-inertial odometry (VIO) system, providing real-time state estimation, alongside velocity aiding from the proprioceptive method DeepVL~\cite{mohit2025DeepVL}, to enable robustness to a lack of visual features in the underwater environment. Among the five cameras, two are used as the front-facing stereo camera pair, with stereo matching performed using \cite{lipson2021raft} for geometric 3D perception.\\
        
        In all robot cases, the proposed unified path-planning framework is also executed onboard the respective autonomy modules to generate motion plans for autonomous operation.

    \subsection{Field Results}            
        To provide a structured overview of the conducted field deployments, Table \ref{tab:field_experiments} summarizes all field experiments performed across three robotic platforms and operating domains. The table reports, for each field environment, the deployed robot platform, key environment characteristics, executed mission behaviors, and associated mission statistics. The table serves as a reference for the detailed qualitative and quantitative results presented in the following subsections.
                
        \subsubsection{\textbf{Underground Mine}}
            The (abandoned) underground mine environment consists of a narrow, tunnel-like structure characterized by constrained cross-sections, uneven surfaces, and multiple branching corridors. The geometry includes long, winding passages with occasional junctions that lead to side branches of varying lengths and visibility. The environment presents limited line-of-sight, visually-degraded conditions, and restricted maneuvering space.
            
            \begin{figure*}[t]
                \centering
                \includegraphics[clip, trim=0cm 0cm 0cm 0cm, width=1\linewidth]{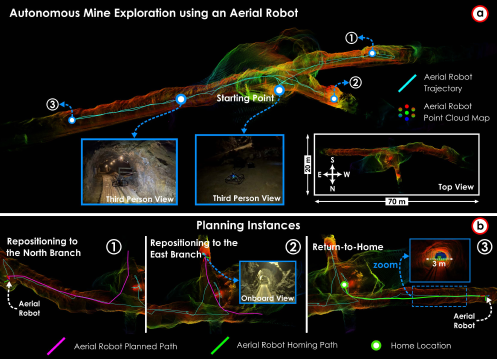}
                \caption{Field results from the \acs{ar2} autonomous aerial robot exploration mission conducted in an underground mine. (a) Full point cloud map generated by the aerial robot, overlaid with the executed robot trajectory and annotated exploration waypoints. Representative third-person views acquired during the mission are shown for reference, along with a top-down view of the reconstructed environment. (b) Key planning instances during the mission, illustrating global repositioning toward unexplored branches and the return-to-home behavior. The planned exploration paths (magenta), homing trajectory (green), and home location are highlighted.}
                \label{fig:mine_exporation_aerial}
            \end{figure*}

            \mypar{Aerial Robot Mission} For the first underground mine mission, the \acs{ar2} aerial robot was deployed from the starting location shown in Fig.~\ref{fig:mine_exporation_aerial}a to autonomously explore the tunnel network. This experiment was intended to stress the planner in a confined, branching, GPS-denied environment, where purely local exploration can terminate prematurely after exhausting the information gain within a single corridor. Starting from the designated location, the robot incrementally expanded the volumetric map while selecting collision-free informative paths through the narrow tunnel geometry. As the mission progressed, the robot encountered multiple branching corridors. When the local planner exhausted the informative viewpoints available within one branch, the global planner used the accumulated topological graph to reposition the robot toward other unexplored branches, as illustrated in Fig.~\ref{fig:mine_exporation_aerial}b.1-b.2. Thus, the key outcome of this deployment is that the robot did not simply follow a single locally informative corridor. It repeatedly recovered from locally completed branches and continued exploration through global repositioning. After the reachable informative regions were explored, the return-to-home behavior was triggered, and the planner generated a homing path back to the starting location, as shown in Fig.~\ref{fig:mine_exporation_aerial}b.3. The resulting three-dimensional map and executed trajectory are shown in Fig.~\ref{fig:mine_exporation_aerial}a. Overall, the robot traversed a total path of $168.5$~m over a mission duration of $4.3$~min.
            \begin{figure*}[t]
                \centering
                \includegraphics[clip, trim=0cm 0cm 0cm 0cm, width=1\linewidth]{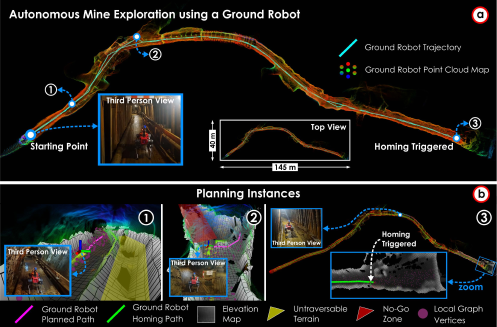}
                \caption{Field results from an autonomous exploration mission conducted in a mine using the \acs{gr1} ground robot with the \acs{am2} payload. (a) Full map generated by the ground platform, alongside its executed trajectory. (b) Planning instances of the local and global planners, including local planning that exploits the elevation map to avoid untraversable and blocked regions, as well as global repositioning toward the home location.}
                \label{fig:mine_exporation_ground}
            \end{figure*}

            \mypar{Ground Robot Mission} For the second underground mine mission, the \acs{gr1} ground robot carrying the \acs{am2} payload was deployed from the mine entrance to perform autonomous volumetric exploration. This experiment complements the aerial mine deployment by evaluating the same exploration behavior under a different definition of motion feasibility: rather than planning freely in 3D space, the ground robot must respect terrain support, elevation-map consistency, slope constraints, and predefined no-go regions. During the mission, the planner selected informative exploration paths while constraining local graph expansion according to the elevation map and restricted areas. As shown in Fig.~\ref{fig:mine_exporation_ground}b.1-b.2, the local planning graph did not expand into no-go regions or areas that were not feasible for ground traversal, while the selected exploration path is shown in pink. This deployment therefore demonstrates that the planning kernel can preserve the same volumetric exploration objective while adapting the admissible graph structure to the ground robot embodiment. After the exploration task was completed, the return-to-home behavior was triggered, and the global planner generated a homing path back to the entrance, shown in Fig.~\ref{fig:mine_exporation_ground}b.3. The complete map generated by the ground platform, together with the executed trajectory, is presented in Fig.~\ref{fig:mine_exporation_ground}a. Overall, the robot traversed a total path length of $357.7$~m over a mission duration of $19.5$~min.

        \subsubsection{\textbf{University Campus}}
            \begin{figure*}[t]
                \centering
                \includegraphics[clip, trim=0cm 0cm 0cm 0cm, width=1\linewidth]{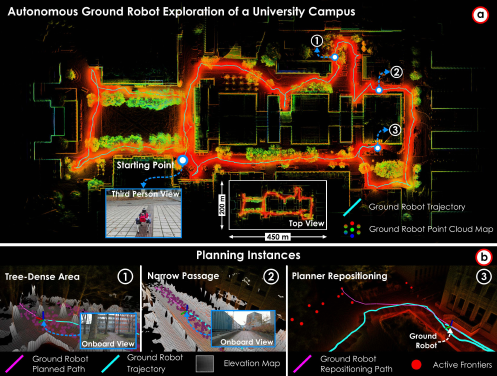}
                \caption{Field results from an autonomous exploration mission conducted on a university campus using the \acs{gr1} ground robot with the sensing and compute configuration of the \acs{am3} payload. (a) Full map generated by the ground platform, alongside its executed trajectory. (b) Planning instances of the local and global planners, including local planning that exploits the elevation map in tree-dense areas and narrow passages, as well as global repositioning toward unexplored regions via frontier selection.}
                \label{fig:campus_exporation}
            \end{figure*}
            This field experiment was conducted in a large-scale outdoor academic environment including long corridors formed by buildings, narrow pedestrian pathways between structures, and tree-dense regions.

            \mypar{Ground Robot Mission} To evaluate long-range autonomous exploration in a realistic outdoor environment, the \acs{gr1} robot was deployed with the sensing and compute payload \acs{am3} on the university campus. This experiment stresses a different aspect of the planner than the underground mine deployments: instead of confined tunnel topology, the robot must operate over a larger spatial scale containing open areas, tree-dense regions, narrow passages, and locally ambiguous routes. Starting from the designated location, the robot incrementally expanded the map by executing informative paths selected by the planner. Fig.~\ref{fig:campus_exporation}b shows three representative planning instances from the mission. In Fig.~\ref{fig:campus_exporation}b.1-b.2, the planner generates collision-free paths through limited-clearance areas and dense obstacle distributions, including tree-dense regions and narrow passages. Fig.~\ref{fig:campus_exporation}b.3 shows a global repositioning maneuver that enabled the robot to escape a locally exhausted region and continue toward unexplored space. The key outcome of this deployment is that the local--global planning structure remained effective over a long-duration outdoor mission, supporting both local obstacle-aware exploration and global repositioning when local progress became limited. The full point cloud map produced during the experiment, together with the executed ground robot trajectory, is presented in Fig.~\ref{fig:campus_exporation}a. Overall, the robot traversed a total path length of $1228.6$~m over a mission duration of $49.5$~min.

        \subsubsection{\textbf{Forest}}
            \begin{figure*}[t]
                \centering
                \includegraphics[clip, trim=0cm 0cm 0cm 0cm, width=1\linewidth]{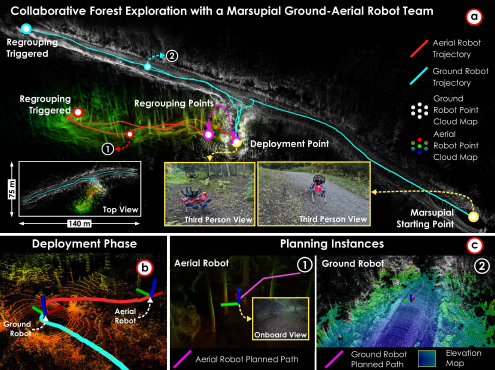}
                \caption{Field results from a collaborative exploration of the marsupial ground-aerial robot team (\acs{ar1} with \acs{am1} payload and \ac{gr1} with \ac{am4} payload) in a forest environment. (a) Combined map generated by both robots, highlighting the start, deployment, and regrouping locations. The aerial robot is deployed when the ground robot cannot continue its mission due to untraversable terrain. (b) Deployment phase of the aerial robot. (c) Planning instances for both robots.}
                \label{fig:forest_exporation}
            \end{figure*}

            \begin{figure*}[t]
                \centering
                \includegraphics[clip, trim=0cm 0cm 0cm 0cm, width=1\linewidth]{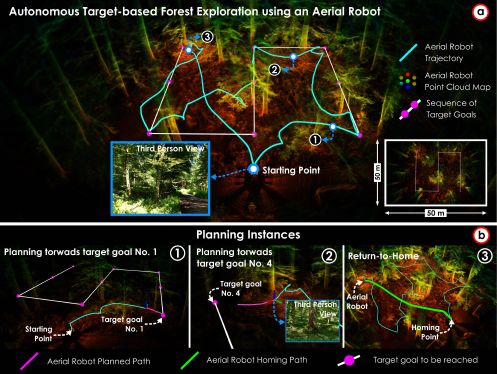}
                \caption{Field results from an autonomous target reach mission conducted in a forest environment using the \acs{ar1} aerial robot. (a) Full point cloud map reconstructed during the mission, overlaid with the executed aerial trajectory and the sequence of target goals. The starting location, intermediate planning instances, and a representative third-person view are indicated, together with a top-down view of the explored area. (b) Key planning instances illustrating motion planning toward successive target goals and the return-to-home behavior. Planned target reach paths (magenta), homing trajectory (green), and target goals are highlighted.}
                \label{fig:forest_target}
            \end{figure*}

            Two field experiments were conducted in forest environments to evaluate the proposed planning framework under different operational settings. Although both experiments were performed in outdoor forested areas, they were carried out at distinct locations with different terrain and vegetation characteristics. 

            \mypar{Marsupial Ground–Aerial Robot Team Mission}
            The first forest experiment was conducted in an environment with uneven terrain and dense vegetation, where some regions were traversable by the ground robot while others required aerial access. This deployment therefore evaluates cross-morphology reuse of the proposed planner within a single heterogeneous mission. A marsupial ground--aerial robot team was used, with the aerial robot initially carried by the ground platform. Multi-robot coordination was handled by a lightweight mission-level layer operating above the planner. Each robot ran an independent instance of \acs{omniplanner} with its corresponding embodiment adaptation layer, while coordination was achieved through deployment decisions, shared map information, and autonomous regrouping commands.

            The mission was initiated from a predefined starting location. During the initial phase, the ground robot explored the environment while carrying the aerial robot and monitored terrain accessibility to determine when aerial deployment was required. Once the ground robot encountered untraversable terrain, the aerial robot was deployed to explore regions beyond the reach of the ground platform. At deployment, the robots established a common reference frame, and the ground robot provided the aerial robot with the relevant map information and an initial target region. After deployment, both robots continued planning independently and exchanged pose-stamped map information whenever communication was available, allowing each robot to update its local volumetric representation.

            Fig.~\ref{fig:forest_exporation}a illustrates the collaborative exploration outcome, showing the trajectories executed by the ground robot (cyan) and aerial robot (red), together with the fused point cloud map generated during the mission. The aerial robot deployment process is shown in Fig.~\ref{fig:forest_exporation}b, while representative planning instances for both robots are presented in Fig.~\ref{fig:forest_exporation}c.1-c.2. These examples highlight how the same planning framework produced terrain-aware paths for the ground robot and collision-free 3D paths for the aerial robot in the same cluttered forest environment. Regrouping events were initiated when the remaining time budget became limited, allowing the robots to reestablish coordination. The key outcome of this experiment is that the planner did not require a separate architecture for the two robots. The same planning kernel was reused across both embodiments, while the mission-level layer handled deployment, map exchange, and regrouping. The \acs{gr1} and \acs{ar1} robots traversed total path lengths of $310.8$~m and $91.3$~m, respectively, over mission durations of $13.9$~min and $2.5$~min.
            
            \mypar{Aerial Robot Mission}
            The second forest experiment was conducted in a different forest environment characterized by tall trees, dense canopy coverage, and cluttered three-dimensional vegetation structures. This setting stresses aerial navigation in limited free space with reduced visibility and complex obstacle distributions. Unlike the previous forest mission, this experiment evaluates the \ac{tr} behavior rather than volumetric exploration. The \acs{ar1} aerial robot was initialized at a predefined starting location, and a sequence of spatial target goals was specified within the forested area. At each planning iteration, the planner generated collision-free trajectories toward the current target while accounting for surrounding vegetation and previously mapped obstacles. After reaching each target, the objective was updated to the next goal in the sequence, as shown in Fig.~\ref{fig:forest_target}a.

            Representative planning instances are shown in Fig.~\ref{fig:forest_target}b.1-b.2, where the planner adaptively generates feasible paths through densely cluttered vegetation. After the target sequence was completed, the return-to-home behavior was triggered, and the planner generated a homing trajectory back to the starting location, as depicted in Fig.~\ref{fig:forest_target}b.3. The key outcome of this deployment is that goal-directed navigation is realized by changing the target position on top of the shared planning kernel. Overall, the robot traversed a total path length of $129.8$~m over a mission duration of $5.1$~min.

        \subsubsection{\textbf{Ballast Water Tank}}
            \begin{figure*}[t]
                \centering
                \includegraphics[clip, trim=0cm 0cm 0cm 0cm, width=1\linewidth]{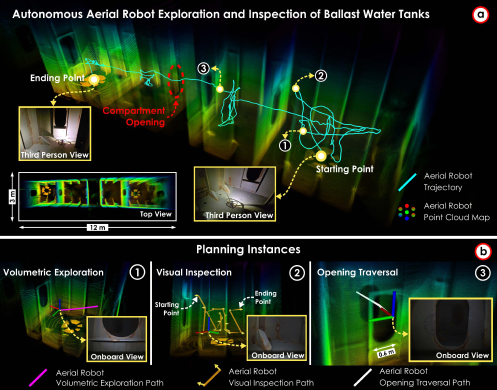}
                \caption{Field results from an autonomous exploration-inspection mission of a ballast water tank using the \acs{ar1} aerial robot. (a) Final map generated by the aerial platform, alongside its executed trajectory. (b) Planning instances of the behaviors executed during the mission, including volumetric exploration using LiDAR point clouds, visual inspection using a camera sensor, and opening traversal based on LiDAR point cloud opening detection.}
                \label{fig:btw_exporation_inspection}
            \end{figure*}
            
            This field experiment was conducted in a confined industrial environment consisting of interconnected ballast water tank compartments with narrow passages and complex internal geometry.  
            
            \mypar{Aerial Robot Mission}
            To evaluate behavior composition in a confined industrial environment, the \acs{ar1} aerial robot was deployed to perform a fully autonomous mission inside the ballast water tank. This experiment differs from the standalone exploration deployments by combining volumetric exploration, visual inspection, and constrained passage traversal within the same mission. Starting from the designated entry point, the robot first explored the interior volume and incrementally built a three-dimensional map of the tank structure. The mapped representation was then used to initiate visual inspection of the observed structural surfaces, enabling close-range sensing without requiring a separate offline inspection pipeline or prior model of the tank. Fig.~\ref{fig:btw_exporation_inspection}a presents the final point cloud map generated during the mission together with the executed aerial robot trajectory.

            During the mission, the robot navigated through multiple compartments and traversed a narrow opening to access adjacent sections of the tank before reaching the designated end point. Fig.~\ref{fig:btw_exporation_inspection}b illustrates representative planning instances corresponding to the different autonomous behaviors executed during the mission, including volumetric exploration for map acquisition, visual inspection for close-range sensing of structural elements, and opening traversal for navigating through constrained passages. The key outcome of this deployment is that exploration and inspection were executed as coupled behaviors on top of the same planning kernel and map representation, rather than as separate task-specific pipelines. Overall, the robot traversed a total path length of $45.1$~m over a mission duration of $4.2$~min.

        \subsubsection{\textbf{Submarine Bunker}}
            \begin{figure*}[t]
                \centering
                \includegraphics[clip, trim=0cm 0cm 0cm 0cm, width=1\linewidth]{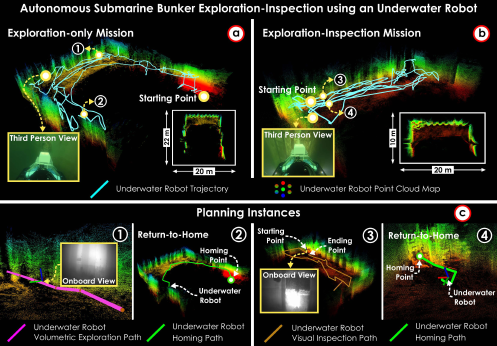}
                \caption{Field results from two autonomous missions conducted in a submarine bunker environment (dry dock) using the \acs{ur1} underwater robot with the \acs{am5} payload: an exploration-only mission and an exploration-inspection mission. (a) Exploration-only mission showing the reconstructed point cloud map and executed trajectory from the deployment location. (b) Exploration-inspection mission, in which VE behavior is followed by the \acs{vi} behavior of selected structural regions. (c) Representative planning instances: (1)-(2) correspond to the exploration-only mission and illustrate volumetric exploration and return-to-home paths, while (3)-(4) correspond to the exploration-inspection mission and show visual inspection path generation followed by the homing path.}
                \label{fig:bunker_exploration_inspection}
            \end{figure*}
            
            The experiments were conducted in a submarine bunker environment. The water exhibited low visibility, resulting in perceptually degraded conditions that challenge both state estimation and collision-aware planning.

            \mypar{Underwater Robot Mission} 
            Two missions were conducted using the \acs{ur1} underwater robot to evaluate the proposed framework under different operational objectives: an exploration-only mission and an exploration--inspection mission. These experiments test the underwater adaptation of the planner, where unconstrained expansion into open water is undesirable because it can lead to uninformative paths away from the infrastructure of interest. Therefore, in both missions, the robot was deployed from a predefined starting location and initialized in \ac{ve} mode, while the local planning graph was constructed by sampling vertices in close proximity to the surrounding structure. This proximity-biased graph construction guided exploration along the bunker geometry and prevented expansion into open-water regions, while still enforcing collision constraints.

            In the exploration-only mission, the robot autonomously explored the environment and generated the three-dimensional point cloud map shown in Fig.~\ref{fig:bunker_exploration_inspection}a. Representative planning instances are shown in Fig.~\ref{fig:bunker_exploration_inspection}c.1-c.2. After the exploration task was completed, the return-to-home behavior was triggered, and the planner generated a homing path back to the deployment location. The key outcome of this mission is that the same planning kernel could be used underwater by modifying the graph admissibility and sampling strategy, via the embodiment adaptation layer,  to reflect the sensing and operational constraints of infrastructure-adjacent underwater navigation. The robot traversed a total path length of $238.9$~m over a mission duration of $13.9$~min.

            In the exploration-inspection mission, the robot first performed volumetric exploration and then used the mapped surface to generate a visual inspection path for detailed observation of structural elements, as shown in Fig.~\ref{fig:bunker_exploration_inspection}b. Since the \ac{ve} behavior in this mission was similar to the exploration-only case, Fig.~\ref{fig:bunker_exploration_inspection}c.3-c.4 focuses on representative planning instances from the \ac{vi} behavior. After the inspection task was completed, the robot autonomously returned to the starting location. This mission complements the ballast water tank deployment by showing that the exploration--inspection sequence is not specific to aerial robots. The same behavior composition transfers to an underwater platform with different sensing, motion, and visibility constraints. The robot traversed a total path length of $156.3$~m over a mission duration of $8.2$~min.

    \subsection{Summary}
        The field experiments evaluate \acs{omniplanner} across eight onboard autonomous deployments spanning underground mines, outdoor campus environments, forests, confined industrial tanks, and underwater infrastructure. These missions exercise different combinations of robot morphology and task behavior, with aerial, ground, and underwater platforms performing volumetric exploration, visual inspection, and target reach. The results show that \acs{omniplanner}’s unified planning kernel can be reused across substantially different robots and tasks, while the adaptation layers and local--global planning structure enable reliable execution under field conditions. Across the deployments, the planner sustained autonomous operation under practical stressors including locally exhausted exploration regions, constrained passages, traversability limits, cluttered vegetation, underwater sensing limitations, and mission-level homing requirements, without requiring separate planning pipelines for each platform or behavior.
        

    \section{DISCUSSION}
\label{sec:discussion}

\subsection{Implications of the Unified Architecture}
    The simulation studies and field deployments presented in this work highlight several implications of using a unified planning architecture across robot morphologies and mission objectives. From a methodological perspective, \acs{omniplanner} indicates that exploration, inspection, and target reach can be formulated around a common local--global planning structure, where task differences are expressed through objective functions rather than through separate planning pipelines. This provides a practical mechanism for reusing the same planning backbone across different autonomous behaviors.

    From an implementation perspective, the unified architecture reduces software fragmentation by avoiding separate planners for each robot type or mission objective. This simplifies development and deployment, since platform-specific requirements are handled through embodiment-dependent sampling, collision checking, and edge validation, while the core planning logic remains unchanged. The same structure also provides a common interface for future heterogeneous multi-robot coordination, where different robot types can operate over compatible planning and map representations.

\subsection{Lessons Learned}

    \mypar{Representation--Task Coupling} The experiments show that reusable planning across domains requires environment representations that support both motion feasibility and task evaluation. The volumetric map provides a common basis for collision checking, frontier extraction, and information gain computation across aerial, ground, and underwater missions. However, additional representations remain necessary when required by the task or embodiment. For example, elevation maps are essential for ground robots to reason about terrain support and slope, while inspection missions require explicit surface and visibility reasoning to evaluate viewpoint quality.

    \mypar{Local--Global Planning Balance} The evaluations and deployments demonstrate that maintaining both local and global planning structures is important in large, cluttered, and partially observed environments. The dense local graph enables short-horizon collision-free motion, immediate viewpoint selection, and fine-grained negotiation of highly cluttered settings and narrow passages. In contrast, the global graph supports long-range repositioning when nearby informative regions are exhausted. This local--global coupling allows the planner to react seamlessly to nearby geometric constraints while still maintaining the ability to escape locally exhausted or dead-end regions. This behavior was particularly important in tunnel-like, branched, and confined environments, where purely local decisions can lead to dead ends or repeated exploration of already observed regions.

    \mypar{Sensing Geometry and Viewpoint Selection} The deployments further highlight that planning performance depends not only on where the robot can move, but also on what the robot can observe from each candidate configuration. FoV limits, sensing range, occlusions, sensor orientation, and camera actuation directly affect the utility of candidate viewpoints. This is especially important for inspection missions and confined environments, where observing the desired surfaces may require deliberate viewpoint selection rather than simply reaching collision-free positions.

    \mypar{Mission-Level Autonomy} The field missions show that practical autonomy requires reliable operation under imperfect mapping, constrained sensing, and changing local geometry. Across underground, industrial, urban, and underwater deployments, the planner repeatedly updated motion feasibility, collision constraints, and information gain using only the currently available map. At the same time, successful operation also depends on mission-level mechanisms beyond the path planner itself, including return-to-home decisions, endurance limits, mission time budgets, and safe termination criteria. These observations reinforce the importance of combining bounded local planning, persistent global connectivity, embodiment-aware validity checks, and mission-level safeguards for robust operation in real field conditions.

\subsection{Theoretical Properties and Practical Guarantees}
   It should be clarified that \acs{omniplanner} does not claim to provide global optimality or completeness guarantees for the full continuous partially observable informative path planning problem. Instead, its guarantees are defined with respect to the finite-resolution map and the graph representation constructed online at planning time. All vertices inserted into the local and global graphs are validated according to the embodiment-specific constraints, including collision clearance for aerial and underwater robots, and traversability, terrain support, and inclination constraints for ground robots. Similarly, graph edges are inserted only when the interpolated motion between two valid vertices remains admissible under the same map, robot-body, and motion-feasibility checks. Therefore, any path returned by the planner is feasible with respect to the current map representation, robot model, graph discretization, and edge-validation resolution.

    For a fixed constructed graph, the graph-search component is complete in the sense that, if a valid graph path exists from the current robot configuration to the selected local goal, frontier, inspection viewpoint, target, or home vertex, the planner will recover such a path. If no such graph path exists, then the planner correctly reports that no solution is available within the current finite graph representation. This property should be interpreted as graph-level completeness rather than completeness in the continuous configuration space. Consequently, if a feasible continuous path is not represented by the current map resolution, sampling density, connection radius, local planning volume, graph-size limits, or edge-validation resolution, the planner may fail to recover it even if such a path exists in the true environment.

    These observations also explain the practical sensitivities observed in the experiments. Sparse or overly restrictive graphs can reduce connectivity and slow exploration progress, while overly dense graphs increase collision-checking, graph-search, and gain-evaluation costs. Edge-checking resolution affects safety near obstacles, while visibility-driven behaviors remain sensitive to FoV, sensing range, occlusions, sensor orientation, and camera actuation limits. For exploration and inspection, the corresponding completion criteria apply only to the observable subset of the environment, since some volume or surface regions may remain unobservable due to sensing limits, occlusions, or infeasible viewpoints, as captured by the residual volume and residual surface definitions in Section~\ref{sec:problem_statement}.

\subsection{Limitations and Future Directions}
    The current framework is primarily applicable to systems that can be modeled as holonomic or near-holonomic at the planning level, including multirotor aerial vehicles, legged and differential-drive ground robots, and hovering underwater vehicles. While this scope covers a broad set of field robotic platforms, it does not directly capture the constraints of strongly non-holonomic systems. For example, fixed-wing aerial vehicles cannot hover and are subject to minimum-turning-radius constraints, which would need to be explicitly represented in the motion model, graph construction process, and trajectory execution layer. More generally, the proposed abstraction is currently applicable when the platform-specific constraints can be represented through sampling and collision checking along bi-directional edges, combined with sensing visibility checks and derivations. When strong non-holonomic constraints, complex whole-body planning (e.g., a legged robot climbing over obstacles), or task-specific perception uncertainty dominate the planning problem, the kernel requires additional modeling rather than only parameter retuning.

    Future work will therefore investigate extensions of the proposed architecture to more restrictive motion models. Additional directions include leveraging the shared planning kernel for coordinated operation of heterogeneous multi-robot teams and integrating perception uncertainty into the sampling of informative viewpoints. This integration would allow the planner to reason not only about expected information gain, but also about the confidence of the underlying environmental representation.    
    
    \section{CONCLUSION}
\label{sec:conclusion}

This paper presented \acs{omniplanner}, a unified planning framework for autonomous exploration, inspection, and target reach missions across diverse aerial, ground, and underwater robotic platforms. The framework is built around a domain-agnostic planning kernel, lightweight embodiment adaptation layers, and modular behavior objectives, enabling the same planning architecture to be reused across multiple robot morphologies and mission types.

The proposed approach was validated through extensive simulation studies and real-world deployments in underground, industrial, terrestrial, and underwater environments. The results demonstrated competitive performance relative to representative state-of-the-art planners while preserving a common architecture across diverse platforms and behaviors. Overall, the presented results show that shared planning principles can be abstracted from robot embodiment and task objective, supporting more transferable and reusable autonomy systems for field robotics.

Future work will extend the framework toward more restrictive motion models, heterogeneous multi-robot coordination, and uncertainty-aware informative viewpoint sampling.


    \section*{NOTATIONS}
    \addcontentsline{toc}{section}{NOTATIONS}
    \glsaddall
    \vspace{-1.4em}
    \printnoidxglossary[title={}]
    
    \bibliographystyle{ieeetr}
    \bibliography{references}

\end{document}